\documentclass[conference]{IEEEtran}
\usepackage{caption}
\usepackage{booktabs}   
\usepackage{graphicx}   
\usepackage{xcolor}     
\usepackage{pifont}     
\newcommand{\tref}[1]{Table~\ref{#1}}
\newcommand{\fref}[1]{Figure~\ref{#1}}

\usepackage{amsmath}   
\usepackage{amssymb}   
\usepackage{pifont}    

\newcommand{\goodmark}{\textcolor{green!80!black}{\ding{51}}} 
\newcommand{\badmark}{\textcolor{red!80!black}{\ding{55}}}   
\newcommand{\Checkmark}{\ding{51}}                           
\newcommand{\XSolidBrush}{\ding{55}}                         
\usepackage[table]{xcolor}   
\usepackage{multirow}        
\usepackage{booktabs}        
\usepackage{colortbl}        

\definecolor{tabfirst}{RGB}{255, 230, 170}   
\definecolor{tabsecond}{RGB}{220, 235, 255}  
\usepackage[numbers]{natbib}
\usepackage{multicol}
\usepackage[bookmarks=true]{hyperref}
\usepackage{caption} 
\usepackage{subcaption}
\usepackage{amsmath,amssymb}
\usepackage{algorithm}
\usepackage{algpseudocode} 

\begin{document}

\title{\LARGE \bf
SuperMap: A Spatio-Temporal SLAM System \\
for Visual-Language Navigation
}

\author{
Shibo Zhao$^{\dagger}$,
Guofei Chen$^{\dagger}$,
Honghao Zhu,
Zhiheng Li,
Changwei Yao, \\
Nader Zantout,
Seungchan Kim,
Wenshan Wang,
Ji Zhang,
and Sebastian Scherer \\
The Robotics Institute, Carnegie Mellon University \\
{\tt\small \{guofeic, shiboz, basti\}@andrew.cmu.edu} \\
$^{\dagger}$Equal contribution
}

\setlength {\marginparwidth }{1cm} 
\pdfoutput=1
\twocolumn[{%
    \renewcommand\twocolumn[1][]{#1}%
    \maketitle 
    \begin{center} 
    \centering 
    \includegraphics[width=0.95\linewidth]{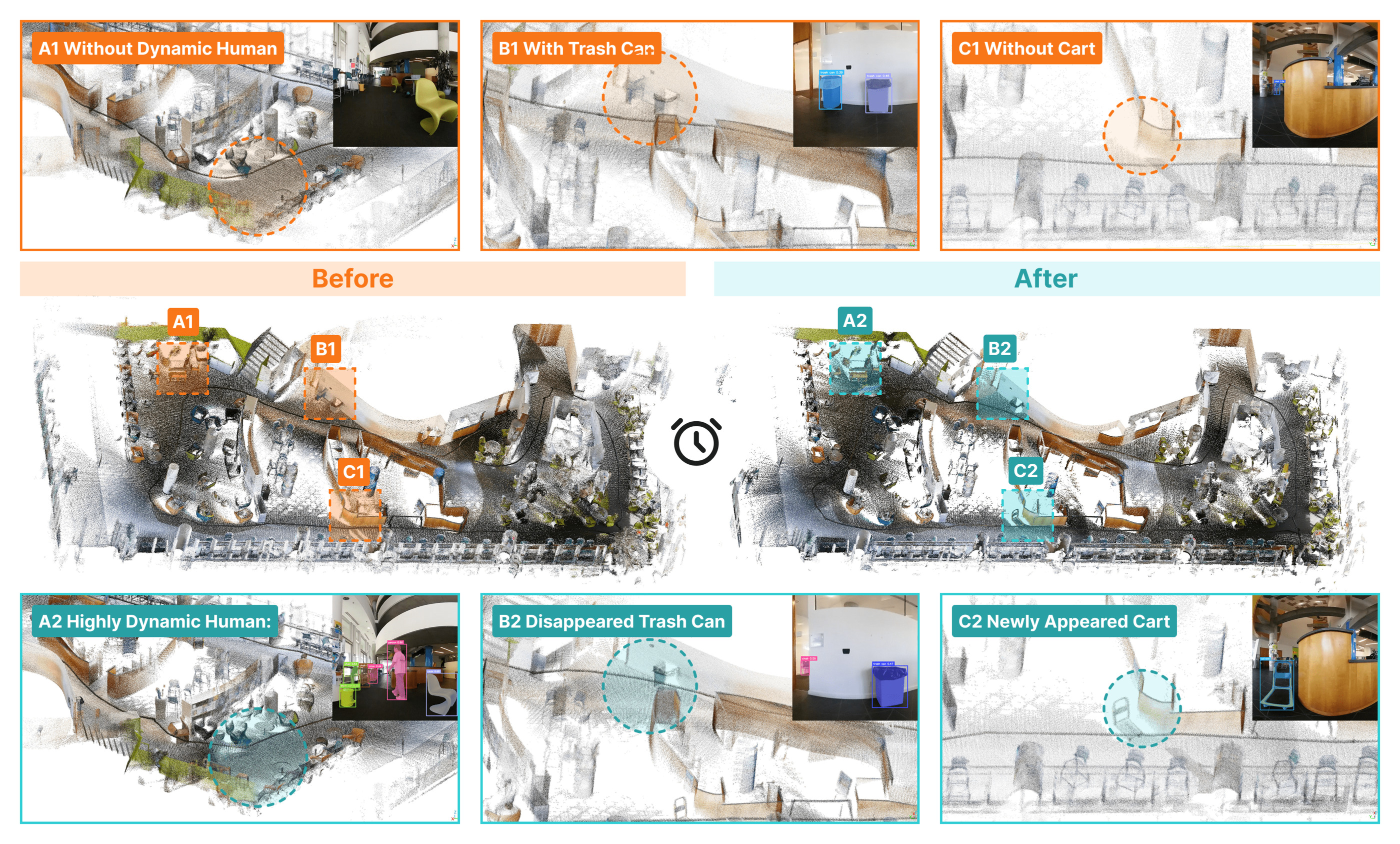}
    \captionsetup{font=small}
    \captionof{figure}{We present SuperMap, a real-time, spatio-temporal SLAM system capable of instance-level object detection, tracking dynamic changes, and maintaining a scene graph for spatial reasoning. The figure shows how SuperMap detects short-term human movements (A1, A2) and long-term environmental changes (B1, B2, C1, C2), such as the disappearance of a trashcan and the appearance of a cart, demonstrating consistent spatio-temporal representation.} 
    \label{fig:key_idea}
    \end{center}%
}]

\begin{abstract}
Robotic navigation in human environments requires a spatio-temporal semantic representation that can reconcile open-vocabulary perception with long-term environmental changes. While foundation models provide strong zero-shot recognition, their predictions are intermittent and view-dependent, and naively integrating them into mapping pipelines leads to identity drift and stale semantics over time. We present SuperMap, a 4D spatio-temporal mapping framework for language-guided navigation that integrates high-frequency geometric SLAM with asynchronous open-vocabulary perception. Our core contribution is a consistency-driven mapping engine that combines 3D-aware instance association/re-activation with a principled existence-and-label confidence update to maintain stable object identities and prune outdated map content under occlusions and scene changes. SuperMap produces a queryable 4D scene-graph representation that interfaces naturally with Vision-Language Models by supporting compositional queries over object semantics, relations, and history. We demonstrate SuperMap on benchmarks and real robots, including dynamic scenes with appearance/disappearance and relocation, and provide ablations and runtime analysis. We release the full system as open-source to provide the community with a deployable baseline for open-vocabulary spatio-temporal mapping. Project website: \color{orange}{\href{https://superodometry.com/supermap}{superodometry.com/supermap}}
\end{abstract}

\IEEEpeerreviewmaketitle

\section{Introduction}

\begin{table*}[t]
\centering
\captionsetup{font=small}
\caption{Capability Checklist for Semantic Mapping and Semantic SLAM Methods. \textcolor{green!80!black}{\Checkmark indicates support}, \textcolor{red!80!black}{\XSolidBrush indicates limited or no support}}
\resizebox{\linewidth}{!}{%
\begin{tabular}{l|cccccc|cccc} 
\toprule
\textbf{Capability} & \multicolumn{6}{c|}{\textbf{Semantic Mapping}} & \multicolumn{4}{c}{\textbf{Semantic SLAM}}\\
\midrule
& HOV-SG\cite{werby2024hierarchical} & ConceptGraphs\cite{gu2024conceptgraphs} & RayFronts\cite{alama2025rayfronts} & OpenScene\cite{peng2023openscene} & OpenMask3D\cite{takmaz2023openmask3d} & SeeGround\cite{li2025seeground} & Kimera\cite{rosinol2020kimera} & OVO-SLAM\cite{martins2024ovo} & Khronos\cite{schmid2024khronos} & SuperMap (Ours)\\ 
\midrule
Real-Time      &            \textcolor{red!80!black}{\XSolidBrush}&               \textcolor{red!80!black}{\XSolidBrush}&             \textcolor{green!80!black}{\Checkmark}&             \textcolor{red!80!black}{\XSolidBrush}&               \textcolor{red!80!black}{\XSolidBrush}&             \textcolor{red!80!black}{\XSolidBrush}&           \textcolor{green!80!black}{\Checkmark}&         \textcolor{green!80!black}{\Checkmark}&         \textcolor{red!80!black}{\XSolidBrush}&  \textcolor{green!80!black}{\Checkmark}\\ 
Open-Vocabulary&            \textcolor{green!80!black}{\Checkmark}&               \textcolor{green!80!black}{\Checkmark}&             \textcolor{green!80!black}{\Checkmark}&             \textcolor{green!80!black}{\Checkmark}&               \textcolor{green!80!black}{\Checkmark}&             \textcolor{green!80!black}{\Checkmark}&           \textcolor{red!80!black}{\XSolidBrush}&         \textcolor{green!80!black}{\Checkmark}&         \textcolor{red!80!black}{\XSolidBrush}&  \textcolor{green!80!black}{\Checkmark}\\ 
Instance-Level &    \textcolor{green!80!black}{\Checkmark}  &  \textcolor{green!80!black}{\Checkmark}   &   \textcolor{red!80!black}{\XSolidBrush}   &  \textcolor{red!80!black}{\XSolidBrush}   &   \textcolor{red!80!black}{\XSolidBrush}   &             \textcolor{red!80!black}{\XSolidBrush}&    \textcolor{red!80!black}{\XSolidBrush}       &     \goodmark    &         \textcolor{red!80!black}{\XSolidBrush}&  \textcolor{green!80!black}{\Checkmark}\\ 
Short-Term  & \badmark & \badmark & \badmark & \badmark & \badmark & \badmark           &  \textcolor{red!80!black}{\XSolidBrush}         &         \textcolor{red!80!black}{\XSolidBrush}&         \textcolor{green!80!black}{\Checkmark}&  \textcolor{green!80!black}{\Checkmark}\\ 
Long-Term & \badmark & \badmark  & \badmark & \badmark &  \badmark & \badmark  & \badmark & \badmark &         \textcolor{green!80!black}{\Checkmark}&  \textcolor{green!80!black}{\Checkmark}\\ 
Scene Graph    &            \textcolor{green!80!black}{\Checkmark}&               \textcolor{green!80!black}{\Checkmark}&             \textcolor{red!80!black}{\XSolidBrush}&             \textcolor{red!80!black}{\XSolidBrush}&               \textcolor{red!80!black}{\XSolidBrush}&             \textcolor{red!80!black}{\XSolidBrush}&           \textcolor{green!80!black}{\Checkmark}&         \textcolor{red!80!black}{\XSolidBrush}&         \textcolor{red!80!black}{\XSolidBrush}&  \textcolor{green!80!black}{\Checkmark}\\ 
\bottomrule
\end{tabular}
}
\vspace{-15.0pt}
\label{tab:capability}
\end{table*}

Robots deployed in human environments are increasingly expected to execute open-vocabulary navigation goals, e.g., “go to the monitor next to the whiteboard” or “return to the chair that was near the plant earlier.” Fulfilling such commands robustly requires more than accurate pose estimation and geometric reconstruction: the robot must maintain an object-centric spatio-temporal map representation that preserves instance identities, semantic attributes, and spatial relations over time, and that remains valid as the environment evolves. This requirement is difficult to meet in practice because real-world scenes are dynamic—objects may be occluded, relocated, removed, or newly introduced during operation. Consequently, systems that rely primarily on per-frame recognition or static semantic maps often degrade over long horizons: instance identities fragment, semantic content becomes stale, and the relational structure needed for language grounding is unavailable or inconsistent.

Recent progress in semantic SLAM and open-vocabulary perception has produced strong components—state estimation and dense reconstruction, 2D tracking, and open-vocabulary detection/segmentation \cite{rosinol2020kimera, krishna20233ds}. However, simply combining these modules rarely yields a map that is reliably usable for navigation. Many systems remain closed-vocabulary and class-level \cite{rosinol2020kimera, krishna20233ds}, limiting instance identity tracking, while dynamic-SLAM methods often focus on short-term motion (e.g., moving people) and overlook long-term changes such as object relocation outside the robot’s view \cite{qiu2022airdos, yu2018ds, song2022dynavins, cui2019sof}. In practice, missed detections and occlusions fragment identities, scene changes introduce stale map content, and most representations lack an efficient query interface for downstream tasks like language-guided navigation. As a result, there remains a gap between “semantic mapping outputs” and a robot-usable map representation that can be queried and acted upon online.

Recent scene graph approaches \cite{gu2024conceptgraphs, werby2024hierarchical, maggio2024clio, rana2023sayplan} incorporate open-vocabulary detectors \cite{ren2024grounded} to support natural language navigation, but they are time consuming (minutes to hours) to construct and not suitable for real-time mapping. Moreover, they struggle to capture dynamic relationships between objects, limiting their ability to model scene evolution.

In summary, existing approaches predominantly address either dynamic change detection (e.g., identifying moving humans or vehicles) or static object recognition (e.g., identifying objects and their relationships). However, \textit{few methods tackle both challenges simultaneously: understanding which objects change and how they evolve spatially and temporally}. Addressing this dual challenge is crucial for robots to effectively perceive and adapt to dynamic real-world environments.

To meet this challenge, we introduce SuperMap: an open-source system that constructs a queryable open-vocabulary 4D scene memory online. We evaluate SuperMap on public benchmarks for open-vocabulary semantic and instance mapping, and on real-robot deployments in dynamic indoor environments that include object appearance/disappearance and long-horizon changes. We provide ablations to attribute mapping improvements to specific maintenance components (association, updates, and graph construction), and runtime profiling to clarify which modules dominate compute. Finally, we demonstrate language-guided robot navigation driven by scene-graph queries, where structured 4D scene memory improves robust grounding compared to using raw detections or video-only inputs.

Our contributions are as follows:

\begin{itemize}
\item \textbf{Open-Vocabulary, Spatio-Temporal Semantic SLAM:}
An online robotic system that builds a persistent, queryable open-vocabulary 4D scene memory suitable for downstream language-conditioned tasks.

\item \textbf{Spatio-Temporal Object Tracking:}
An online pipeline that integrates 2D–3D association, validation,
and change-aware updates to maintain instance consistency under occlusions, partial observations, label variability, and scene change. 

\item \textbf{Instance-level Scene Graph:}
Our 4D scene graph seamlessly incorporates spatial and temporal information for each object, equipping robots with instance-level spatio-temporal reasoning capabilities (e.g., locating moved objects, recalling past scenes).

\item \textbf{Open-Source Framework:}
We will release our change-detection benchmark, comprehensive ablations and runtime profiling, and the real-robot visual–language navigation pipeline to facilitate reproducible research in the robotics community.
\end{itemize}

\begin{figure*}[h!]
    \centering
    \includegraphics[width=1.0\linewidth]{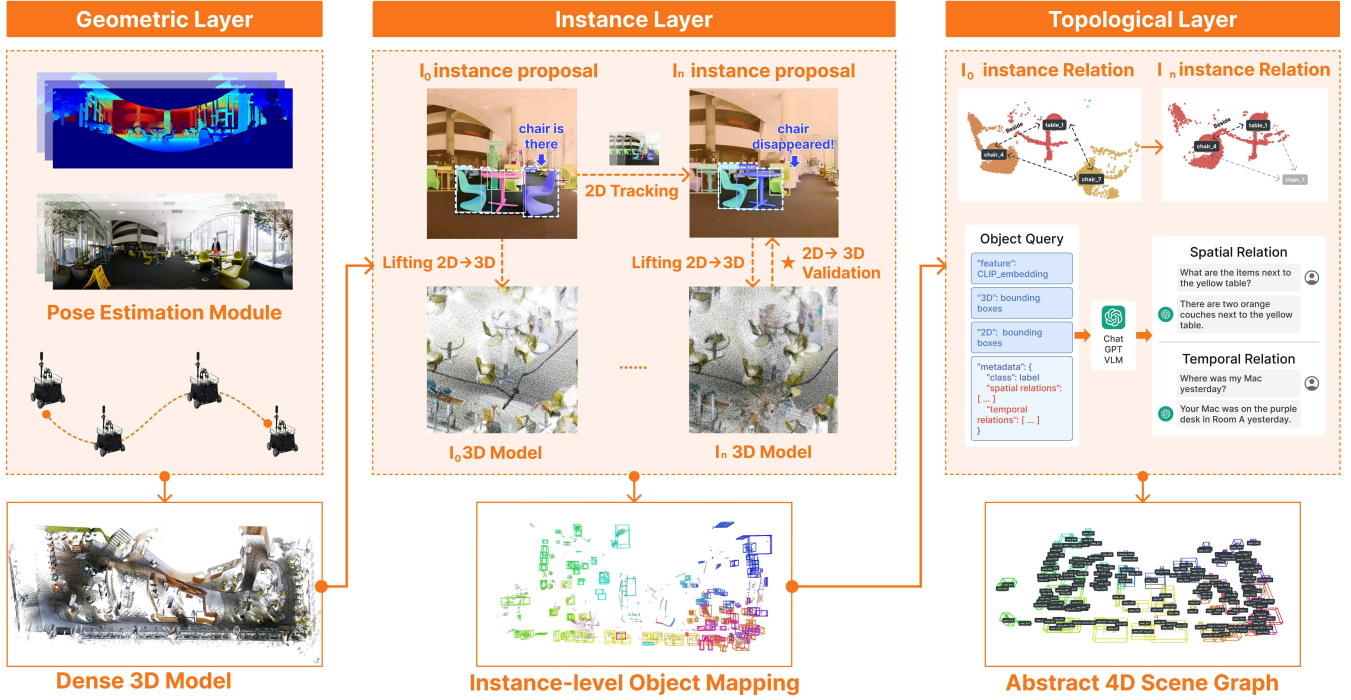}
    \captionsetup{font=small}
    \caption{\textbf{Overview}. The system delivers accurate state estimation together with a rich, multi-level map representation that spans from geometric details to abstract semantics. It consists of three main modules. First, an online 3D reconstruction module built on a LiDAR SLAM backbone performs state estimation and generates a dense 3D model of the environment. Next, the estimated state and reconstructed geometry are used to identify short-term and long-term object changes, enabling instance-level object mapping. Finally, the object-level representation is organized into an abstract 4D scene graph that captures both spatial and temporal relationships among objects.}
    \label{fig:method}
    \vspace{-5pt}
\end{figure*}

\section{Related Works}
\subsection{Semantic Mapping}
Equipping mobile robots with the ability to perceive and understand their surroundings is crucial for real-world applications. To this end, researchers have developed semantic maps to capture environmental meaning and context~\cite{raychaudhuri2025semantic}. However, many recent methods~\cite{peng2023openscene,gu2024conceptgraphs} rely on offline processing, requiring pre-collected environmental data (images and depth) to build semantically labeled maps shown in \tref{tab:capability}.

For example, OpenScene~\cite{peng2023openscene} uses OpenSeg~\cite{yuan2021hrformer} to extract CLIP embeddings~\cite{radford2021learning} from images and then trains a scene-specific 3D model, but it operates offline, requiring a full scene scan. OpenMask3D~\cite{takmaz2023openmask3d} back-projects 2D object masks from multiple views to form 3D segments, while SeeGround~\cite{li2025seeground} combines 2D-VLM with 3D geometric cues for instance segmentation. Despite improving 3D accuracy, these methods assume complete scene data, limiting online usability.

LERF~\cite{kerr2023lerf} and LangSplat~\cite{qin2024langsplat} employ radiance fields to build 3D maps with CLIP-based features, offering flexible semantic queries but lacking real-time capability due to computational demands. RayFronts~\cite{alama2025rayfronts} enables online semantic mapping for outdoor environments but struggles with change detection and fails to achieve instance-level 3D semantics.

\subsection{Semantic SLAM} To jointly estimate poses within the pipeline, we transform the mapping pipeline into a SLAM pipeline. In recent years, numerous SLAM systems have integrated semantic information during the mapping process. One notable example is Kimera \cite{rosinol2020kimera}, which offers a real-time semantic mapping solution by building a 3D scene graph. However, it relies on closed-set CNN segmentation, which limits its ability to generalize to novel classes. LiDAR-based SLAM approaches, such as SuMa++ \cite{chen2019suma++} and LIOM \cite{zhao2019robust}, incorporate CNN-based per-point labels into LiDAR maps. Despite their effectiveness, these methods do not support open-vocabulary segmentation, restricting their adaptability to diverse environments. SlideSLAM \cite{liu2024slideslam} and OVO-SLAM \cite{martins2024ovo} employ open-vocabulary detectors for instance segmentation, allowing for more flexible scene understanding. However, these methods face challenges when dealing with \textbf{long-term and short-term dynamic objects}.

\subsection{Open-Vocabulary Scene Graph}
HOV-SG \cite{werby2024hierarchical}, ConceptGraphs \cite{gu2024conceptgraphs} and CLIO \cite{maggio2024clio} segment and label objects from a full point cloud using SAM and CLIP, organizing them into a hierarchical graph with open-ended labels from CLIP. However, HOV-SG and ConceptGraphs require offline 3D reconstruction before applying SAM and CLIP, making them unsuitable for online SLAM or real-time dynamic tasks. CLIO implements real-time open-vocabulary scene graph generation but lacks spatio-temporal capabilities.  

\subsection{Spatio-Temporal Semantic SLAM}
Recently, Khronos \cite{schmid2024khronos} addresses both short-term and long-term dynamics in map representation. However, this method is limited to closed-set objects and lacks the capability for instance-level tracking. Additionally, Khronos struggles to operate in real time, making it unsuitable for robotic operations.
To the best of our knowledge, SuperMap is the first real-time, spatio-temporal, open-vocabulary, and instance-level object mapping framework that effectively handles both short-term and long-term dynamic objects.

\section{Spatio-Temporal SLAM Problem}
\label{sec:problem}

The spatio-temporal SLAM problem aims to provide an explicit scene representation that captures how object-level changes evolve across space and time through incremental data acquisition. We model the scene as a composition of objects, represented as a scene graph where nodes correspond to objects and edges denote spatio-temporal relationships.

In our setting, the system processes a continuous stream of either an RGB-D video sequence or a point-cloud video sequence, where RGB frames
$C_{1:T} = \{C_1, C_2, \dots, C_T\}$ and depth(LiDAR) frames
$D_{1:T} = \{D_1, D_2, \dots, D_T\}$ are given as input, and $T$ is the sequence length and $t \in \{1,\dots,T\}$ is the time index. Camera poses
$P_{1:T} = \{P_1, P_2, \dots, P_T\}$ must be estimated. At each time $t$, the system outputs a consistent global map $\mathcal{M}_t$ consisting of semantic instance-level objects
$\mathcal{O}_t = \{\mathcal{O}_t^{\,j}\}_{j=1}^{N_t}$, where $j$ indexes objects and $N_t$ is the number of objects at time $t$.

The spatio-temporal SLAM problem is defined as follows: given the current observation
$Q_t = \{C_t, D_t\}$ and the existing map $\mathcal{M}_{t-1}$, estimate pose $P_t$, determine the instance IDs $\mathcal{I}_t$ and the updated map $\mathcal{M}_t$ by maximizing (or estimating)
\begin{equation}
P(\mathcal{I}_t, \mathcal{M}_t, P_t \mid \mathcal{M}_{t-1}, Q_t).
\end{equation}
Applying the chain rule, we obtain
\begin{equation}
\begin{aligned}
P(\mathcal{I}_t, \mathcal{M}_t, P_t \mid \mathcal{M}_{t-1}, Q_t)
&=
\underbrace{P(P_t \mid \mathcal{M}_{t-1}, Q^{\mathrm{}}_t)}_{\text{pose estimation (geometric)}} \\
&\quad \times
\underbrace{P(\mathcal{I}_t \mid \mathcal{M}_{t-1}, Q_t, P_t)}_{\text{spatio-temporal instance association}} \\
&\quad \times
\underbrace{P(\mathcal{M}_t \mid \mathcal{M}_{t-1}, Q_t, P_t, \mathcal{I}_t)}_{\text{online map update}}.
\end{aligned}
\label{eq:problem_factorization}
\end{equation}
This involves three tasks: pose estimation, spatio-temporal object instance association and the online map update~\cite{deng2025openvox}.

\section{Method}
\label{sec:method}

The architecture of our spatio-temporal SLAM system, SuperMap, is shown in~\fref{fig:method}. Given synchronized depth measurements, IMU data, and RGB images, SuperMap runs fully onboard to jointly estimate the robot state and maintain an instance-level spatio-temporal map representation as the environment evolves.  We describe three main components: (1) \textit{Online 3D Reconstruction} for dense geometric mapping, (2) \textit{Spatio-Temporal Object Updates} for robust instance association, and (3) \textit{Spatio-Temporal Scene Graph Construction} that exposes the map as a queryable interface for VLN.

\subsection{Geometric Layer: Online 3D Reconstruction}
We use SuperOdometry~\cite{zhao2021super} to obtain the robot pose $P_t$ and a colorized dense 3D reconstruction from image, depth/LiDAR, and IMU streams. The observation $Q_t$ requires accurate camera poses $P_t$ and depth observations $D_t$.  For each time $t$, we estimate the robot's state in $SE(3)$, denoted as $\mathbf{T}_{WB}^{(t)}$, where $W$ is the world frame and $B$ is the body frame. Given a fixed extrinsic calibration $\mathbf{T}_{BC}$, the camera pose $P_t$ is derived as:
\begin{equation}
P_t = \mathbf{T}_{WC}^{(t)} = \mathbf{T}_{WB}^{(t)} \cdot \mathbf{T}_{BC}
\end{equation}

This estimated pose and the synchronized depth $D_t$ form the observation $Q_t$. These geometric priors are critical for: (i) back-projecting 2D semantic observations into the 3D world, (ii) performing motion compensation for temporal tracking, and (iii) maintaining global map consistency. By anchoring all subsequent instance-level detections to this metric foundation, we ensure that the resulting spatio-temporal map is physically grounded and suitable for downstream navigation tasks.

\subsection{Instance Layer: Spatio-Temporal Instance Association}
For each image frame we first infer the 2D instance-level detection and segmentations using vision models \cite{liu2024grounding} \cite{ravi2024sam}, then address the estimation of instance IDs $\mathcal{I}_t$ by solving $P(\mathcal{I}_t \mid \mathcal{M}_{t-1}, Q_t)$. The objective is to assign a unique instance ID to 2D detections by leveraging 3D spatial consistency.

\subsubsection{Hybrid Tracking State}
We define the tracklet state $\mathbf{S}_i(t) \in \mathbb{R}^6$ (or $\mathbb{R}^8$ if including scale velocity) as a decoupled vector:
\begin{equation}
\mathbf{S}_i(t) = [\mathbf{c}_i(t)^\top, \mathbf{s}_i(t)^\top, \dot{\mathbf{c}}_i(t)^\top]^\top
\end{equation}
where $\mathbf{c}_i(t) = [x, y]^\top$ is the 2D image centroid, $\mathbf{s}_i(t) = [w, h]^\top$ represents the bounding box width and height, and $\dot{\mathbf{c}}_i(t)$ is the translational velocity in the image plane.

\subsubsection{3D-to-2D Motion-Compensated State}
Since standard 2D trackers (e.g., ByteTrack~\cite{zhang2022bytetrack}) struggle with rapid ego-motion. We resolve this by defining a hybrid tracking state $\mathbf{S}_i(t)$ that is updated via 3D-to-2D Motion Compensation.

Instead of a linear motion model, we derive the predicted centroid $\hat{\mathbf{c}}_i(t)$ by projecting the 3D centroid $\mathbf{X}_i$ of the instance in $\mathcal{M}_{t-1}$ using the current pose $P_t$:

\begin{equation}
\hat{\mathbf{c}}_i(t) = \pi(\mathbf{K} \cdot P_t^{-1} \cdot \mathbf{X}_i)
\end{equation}
The Kalman Filter state transition uses this projection as the prior:
\begin{equation}
\hat{\mathbf{S}}_i(t) = \mathbf{F} \mathbf{S}_i(t-1) + \mathbf{w}_t, \quad \mathbf{w}_t \sim \mathcal{N}(0, \mathbf{Q})
\end{equation}

where $\mathbf{F}$ is the state transition matrix and $\mathbf{w}_t$ represents process noise. This projection-based prior allows the system to maintain identity through occlusions and aggressive robot maneuvers.


\subsubsection{Probabilistic Geometric Consistency Update}
The observation $Q_t$ updates the map $\mathcal{M}_t$ by evaluating the posterior $P(\mathcal{M}_t \mid \mathcal{M}_{t-1}, Q_t, \mathcal{I}_t)$. We model the depth observation for each point $\mathbf{X}_k \in \mathcal{M}_{t-1}$ as a random variable. 

Let $d_{proj} = \|\mathbf{T}_{CW} \mathbf{X}_k\|_z$ be the expected (projected) depth of the point in the current camera frame, and $D(\mathbf{u})$ be the raw sensor depth at the projected pixel $\mathbf{u} = \pi(\mathbf{X}_k)$. The depth residual is defined as:
\begin{equation}
\Delta d = d_{proj} - D(\mathbf{u})
\label{eq:consistency}
\end{equation}

We model the measurement likelihood under the assumption of Gaussian sensor noise, $p(\Delta d) \sim \mathcal{N}(0, \sigma^2)$. The occupancy state $o_k$ of the map point is updated via a log-odds formulation to ensure numerical stability and recursive update capability:
\begin{equation}
L(o_k \mid Q_{1:t}) = L(o_k \mid Q_{1:t-1}) + \text{logit}(P(o_k \mid Q_t))
\end{equation}
where $L(n) = \log \frac{n}{1-n}$. The inverse sensor model $P(o_k \mid Q_t)$ is determined by the thresholded geometric consistency of $\Delta d$:
\begin{equation}
s_k^{(t)} = 
\begin{cases} 
\text{Observable} & \text{if } |\Delta d| \leq \tau_{\epsilon} \\
\text{Unobservable} & \text{if } \Delta d > \tau_{\epsilon} \quad (\text{Point behind surface}) \\
\text{Disappeared} & \text{if } \Delta d < -\tau_{\epsilon} \quad (\text{Point in front of surface})
\end{cases}
\end{equation}
Points classified as dynamic are penalized in the log-odds update, effectively pruning transient or moved objects from the global map to maintain a consistent representation of the environment. This enables the change detection in dynamic scenes, and also provides geometric cue for fusing the semantic inference across frames.


\subsubsection{Bayesian Semantic Fusion}
To handle the uncertainty in the output of vision models, each instance maintains a categorical distribution $P(L_j = c)$. Given observations $z_{1:t}$ from an open-vocabulary detector, the posterior belief is:
\begin{equation}
P(L_j = c \mid z_{1:t}) = \eta \cdot P(z_t \mid L_j = c) \cdot P(L_j = c \mid z_{1:t-1})
\end{equation}
where $P(z_t \mid L_j = c)$ is the detector's confusion matrix. 

For observable points in (9), we update the semantic label using (10), and consequently remove object points whose posterior belief is too small. This fusion suppresses transient misclassifications.

\subsection{ Topological Layer: Spatio-Temporal Scene Graph}

The map $\mathcal{M}_t$ is abstracted into a graph $\mathcal{G} = (\mathcal{V}, \mathcal{E}_s, \mathcal{E}_t)$. Nodes $\mathcal{V}$ represent object instances.
\begin{itemize}
    \item \textbf{Spatial Edges ($\mathcal{E}_s$):} Established via class-dependent geometric predicates. For objects $A, B$:
    $\text{On}(A, B) \iff (z_{min}^A \approx z_{max}^B) \land (\text{IoU}_{xy}(\mathcal{B}_A, \mathcal{B}_B) > \gamma)$.
    \item \textbf{Temporal Edges ($\mathcal{E}_t$):} Link nodes across time based on $\mathcal{I}_t$ results, tracing object trajectories.
\end{itemize}

For real-time operation, we first cluster objects based on their centroid distances and then add a spatial edge when an object pair satisfies class-dependent geometric predicates (e.g., on, beside, under). The temporal edges trace an object’s trajectory from our object association result, preserving its identity throughout the sequence. The resulting graph compactly captures both instantaneous layout and long-term dynamics, providing an abstract representation of the scene that is easy to reason on. Full construction details appear in the supplementary material.

\subsection{Visual Language Navigation via 4D Scene Graphs}
The 4D scene graph $\mathcal{G}$ provides a high-level interface between raw sensor data and the symbolic inputs required by Vision--Language Models (VLMs). By delegating geometric detail to the spatio-temporal map, the VLM can perform zero-shot reasoning over long-term scene dynamics without processing raw point clouds or video streams.

\paragraph{Spatio-Temporal Serialization}
We serialize a local subgraph into a structured text representation for linguistic reasoning. Each node $v \in V$ is encoded by its instance ID, semantic label, and 3D centroid $\mathbf{X}_v \in \mathbb{R}^3$. Spatial and temporal edges ($E_S$, $E_T$) are represented as symbolic relations (e.g., \textit{Instance\_1 on Instance\_2}). Reliability is improved by (1) prepending a schema that defines the graph structure and coordinate frame, and (2) injecting explicit spatio-temporal cues that encourage trajectory- and relation-aware reasoning.

\paragraph{Grounded Actuation and Waypoint Generation}
To convert VLM outputs into deterministic control signals, the model must return target instance IDs within \texttt{<answer></answer>} tags. A parser extracts these IDs and retrieves their 3D centroids $\mathbf{X}_j$ from the map $\mathcal{M}_t$, which are then used as navigation waypoints. This process translates high-level language instructions into precise, grounded robot actions.

\section{Experiments} 
\subsection{Experiment Settings}
\label{sec:experiment_settings}

\begin{table*}[t]
\centering
\captionsetup{font=small}
\caption{\textbf{Class-level Segmentation Benchmarking on ScanNet.} Best results are highlighted as \colorbox{tabfirst}{\textbf{first}} and \colorbox{tabsecond}{second}.}
\resizebox{0.8\linewidth}{!}{%
\begin{tabular}{l|l|ccc|ccc}
\toprule
\textbf{Method} & \textbf{Mapping Approach} & \multicolumn{3}{c|}{\textbf{Without Background}} & \multicolumn{3}{c}{\textbf{With Background}} \\
\cmidrule(lr){3-5} \cmidrule(lr){6-8}
& & mIoU (\%) & f-mIoU (\%) & Acc (\%) & mIoU (\%) & f-mIoU (\%) & Acc (\%) \\
\midrule
ConceptFusion \cite{jatavallabhula2023conceptfusion} & \multirow{4}{*}{point-feature} & 21.76 & 26.71 & 34.10 & 18.57 & 23.06 & 28.77 \\
NACLIP-3D \cite{hajimiri2025pay} & & 22.32 & 24.32 & 33.46 & 22.32 & 24.32 & 33.46 \\
Trident-3D \cite{shi2024harnessing} & & \cellcolor{tabsecond}29.97 & 37.62 & 51.06 & \cellcolor{tabsecond}24.80 & 27.77 & 38.43 \\
RayFronts \cite{alama2025rayfronts} & & \cellcolor{tabfirst}\textbf{41.29} & \cellcolor{tabfirst}\textbf{46.42} & \cellcolor{tabfirst}\textbf{56.76} & \cellcolor{tabfirst}\textbf{32.29} & \cellcolor{tabfirst}\textbf{39.04} & \cellcolor{tabfirst}\textbf{49.15} \\
\midrule
ConceptGraphs \cite{gu2024conceptgraphs} & \multirow{4}{*}{object-level} & 21.62 & 24.32 & 31.05 & 20.83 & 23.61 & 35.80 \\
HOV-SG \cite{werby2024hierarchical} & & 26.79 & 36.05 & 35.17 & 23.48 & 28.92 & \cellcolor{tabsecond}38.52 \\
\textbf{SuperMap (Ours)} & & 27.42 & \cellcolor{tabsecond}43.50 & \cellcolor{tabsecond}55.48 & 22.61 & \cellcolor{tabsecond}29.10 & 33.00 \\
\bottomrule
\end{tabular}%
}
\vspace{-5pt}
\label{tab:class-semantic}
\end{table*}

\begin{table*}[t]
\centering
\setlength{\tabcolsep}{4pt} 
\captionsetup{font=small}
\caption{\textbf{Instance-level Segmentation Benchmark on ScanNet.} Best results are highlighted as \colorbox{tabfirst}{\textbf{first}} and \colorbox{tabsecond}{second}.}
\resizebox{0.8\linewidth}{!}{%
\begin{tabular}{l|cc|cc|cc|cc|cc}
\toprule
\textbf{Method} & \multicolumn{2}{c|}{\textbf{Chair}} & \multicolumn{2}{c|}{\textbf{Window}} & \multicolumn{2}{c|}{\textbf{Refrigerator}} & \multicolumn{2}{c|}{\textbf{Sofa}} & \multicolumn{2}{c}{\textbf{Door}} \\
\cmidrule(lr){2-3} \cmidrule(lr){4-5} \cmidrule(lr){6-7} \cmidrule(lr){8-9} \cmidrule(lr){10-11}
& mAP$_{50}$ & mAP$_{25}$ & mAP$_{50}$ & mAP$_{25}$ & mAP$_{50}$ & mAP$_{25}$ & mAP$_{50}$ & mAP$_{25}$ & mAP$_{50}$ & mAP$_{25}$ \\
\midrule
HOV-SG \cite{werby2024hierarchical} & \cellcolor{tabsecond}4.58 & \cellcolor{tabsecond}4.73 & 0.00 & 0.00 & 0.00 & 0.00 & \cellcolor{tabsecond}30.00 & \cellcolor{tabsecond}31.25 & \cellcolor{tabsecond}9.70 & \cellcolor{tabsecond}10.40 \\
ConceptGraphs \cite{gu2024conceptgraphs} & 0.00 & 2.33 & 0.00 & 0.00 & 0.00 & 0.00 & 0.00 & 0.00 & 0.00 & 0.00 \\
\textbf{SuperMap (Ours)} & \cellcolor{tabfirst}\textbf{63.76} & \cellcolor{tabfirst}\textbf{74.72} & \cellcolor{tabfirst}\textbf{42.20} & \cellcolor{tabfirst}\textbf{67.92} & \cellcolor{tabfirst}\textbf{62.50} & \cellcolor{tabfirst}\textbf{62.50} & \cellcolor{tabfirst}\textbf{33.35} & \cellcolor{tabfirst}\textbf{83.35} & \cellcolor{tabfirst}\textbf{10.00} & \cellcolor{tabfirst}\textbf{25.00} \\
\bottomrule
\end{tabular}%
}
\label{tab:instance-semantics}
\end{table*}

\paragraph{\textbf{Hardware Configuration}}We evaluate SuperMap on a custom-built mecanum-wheeled mobile platform designed for agile indoor navigation. The sensor suite comprises a Livox Mid-360 LiDAR, which provides a 360$^\circ \times$ 59$^\circ$ field of view, and a panoramic camera providing 360$^\circ$ RGB coverage. We implement a software-based timestamp alignment to associate visual and LiDAR streams. All computations, including LVI-odometry, instance tracking, 4D graph maintenance, and VLM interfacing, run onboard in real-time on an Intel i9-14900H CPU and an NVIDIA RTX 4090 Laptop GPU (16GB VRAM). We aim to address the following research questions through our experiments:

\begin{itemize}
    \item \textbf{Semantic Quality:} How does SuperMap’s incremental object-centric mapping compare to state-of-the-art offline and online 3D semantic mapping approaches on the ScanNet benchmark (Sec.~\ref{sec:segmentation})?

    \item \textbf{Spatio-Temporal Consistency:} Can the 4D graph accurately detect scene changes (appearances/disappearances) while maintaining persistent identities for static objects over long-duration trajectories (Sec.~\ref{sec:4D-scene-graph})? 
    
    \item \textbf{Grounded Reasoning:} Does the structured scene graph representation improve the accuracy and token efficiency of VLM reasoning comparing to video input  (Sec.~\ref{sec:reasoning})?

    \item \textbf{Online Visual-language Navigation:} Can the integrated pipeline support end-to-end, zero-shot, language-guided navigation using solely onboard compute (Sec.~\ref{sec:vln})?
\end{itemize}

\subsection{Semantic Quality: Quantitative Results for Class-level and Instance-level Segmentation}
\label{sec:segmentation}
To evaluate the semantic fidelity of SuperMap, we conduct a comparative analysis against state-of-the-art open-vocabulary mapping frameworks, focusing on both class-level precision and instance-level consistency.

\paragraph{Class-level Segmentation}

Table~\ref{tab:class-semantic} summarizes the performance on the ScanNet benchmark. SuperMap achieves a competitive accuracy of \textbf{55.48\%}, outperforming established object-centric baselines such as \textit{ConceptGraphs}~\cite{gu2024conceptgraphs} and \textit{ConceptFusion}~\cite{jatavallabhula2023conceptfusion}. While our accuracy is slightly lower (1.32\%) than the point-feature fusion method \textit{RayFronts}~\cite{alama2025rayfronts}, SuperMap maintains a significantly lower computational footprint, making it more suitable for high-rate robotic feedback loops. These results demonstrate its effectiveness in open-vocabulary segmentation among works on object-centric mapping.

\paragraph{Instance-level Segmentation}
Unlike offline frameworks that rely on global scene context, SuperMap targets online robotic deployment, where incremental data fusion and zero-shot generalization are essential. This setting requires establishing object correspondences from partial, streaming observations in real time. To enable zero-shot perception, we integrate GroundingDINO~\cite{liu2024grounding} for language-grounded detection and SAM2~\cite{ravi2024sam} for robust instance segmentation. We report comparative results against \textit{HOV-SG}~\cite{werby2024hierarchical} and \textit{ConceptGraphs}~\cite{gu2024conceptgraphs} in Table~\ref{tab:instance-semantics}. SuperMap significantly outperforms these baselines, particularly for discrete objects contained within the camera's instantaneous field of view. This gain demonstrates the efficacy of our 3D-aware tracking-by-detection approach over methods relying on point-feature clustering or over-segmented geometries. 


\begin{figure*}[htbp]
    \centering
    \includegraphics[width=1.0\linewidth]{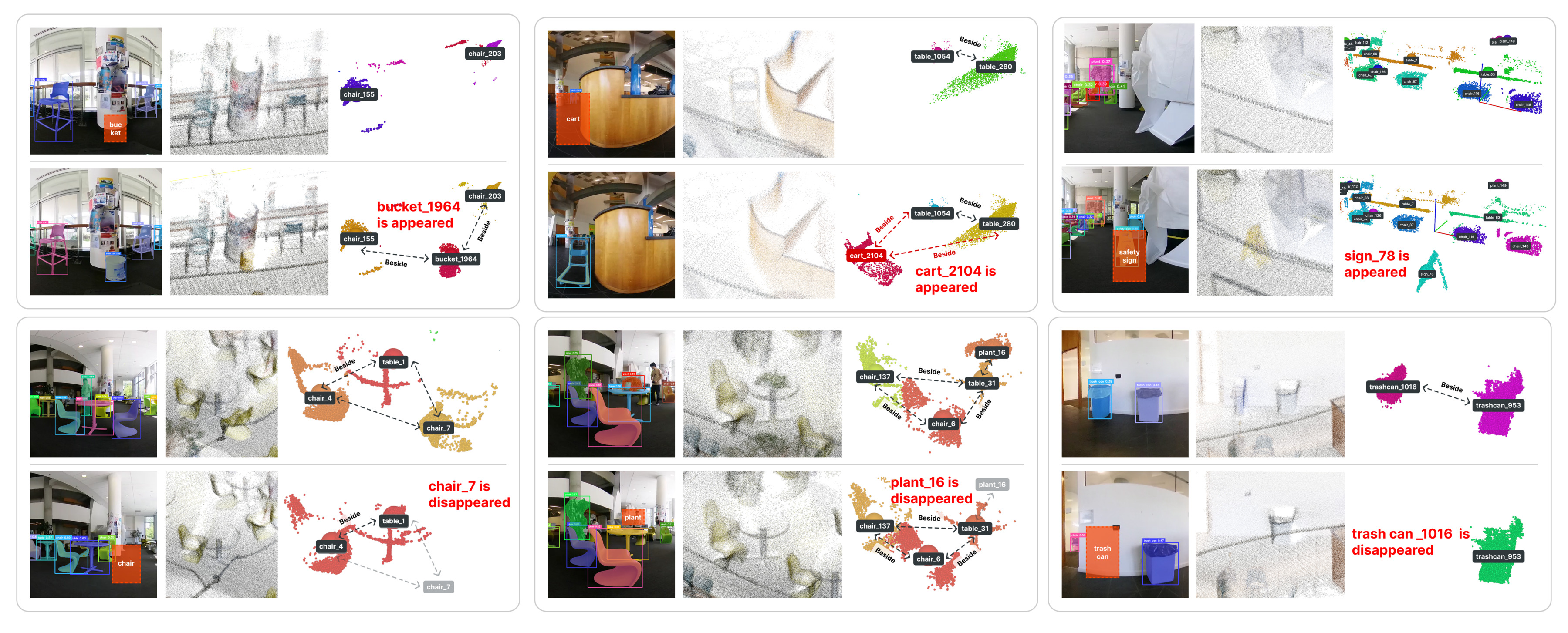}
    \captionsetup{font=small}
    \caption{\textbf{Spatio-temporal Consistency.} For each example (left to right), we show the 2D detections, the dense 3D reconstruction, and the object-centric map rendered in Rerun \emph{before} and \emph{after} the scene change. \textbf{Top:} three appearance events (bucket, cart, safety sign), where newly introduced objects receive new IDs while nearby objects retain their original IDs, demonstrating long-horizon re-identification and stable instance association in 3D. \textbf{Bottom:} three disappearance events (chair, plant, trash can), where the same instance ID is preserved across time despite the object leaving the field of view. 
    }
    
    \label{fig:temporal_consistency}
    \vspace{-10.0pt}
\end{figure*}

\subsection{Spatio-Temporal Consistency on Qualitative Evaluation}
\label{sec:4D-scene-graph}

We evaluate SuperMap’s ability to maintain map integrity over extended durations while identifying environmental changes. In a 10-minute real-world experiment (30m $\times$ 20m indoor area), we introduced structural changes: three objects were removed (a plant, a trash can, and a chair) and three were added (a bucket, a cart, and a safety sign).

\paragraph{Qualitative Performance}
Fig.~\ref{fig:temporal_consistency} qualitatively demonstrates SuperMap’s long-horizon instance consistency under both disappearance and appearance events. The top row shows three objects that disappear from the scene. In all cases, the instance IDs remain consistent over time, indicating that the system preserves object identity even when objects are no longer observed.

For example, the top row shows three objects that newly appear in the scene. In the top-left example, we place a new bucket instance (\textbf{bucket 1964}) between \textbf{Chair 155} and \textbf{Chair 203}. In the bottom-left example, \textbf{Chair 7} disappears between \textbf{table 1} and \textbf{chair 4}. Nonetheless, the small ID label in the Rerun visualization remains unchanged, showing that the system maintains the same instance association before and after the disappearance event.
It demonstrates consistent object tracking and identity maintenance over long time horizons.

\subsection{Spatial-Temporal Consistency on Quantitative Evaluation}

We quantitatively assess SuperMap using six target objects (Fig.~\ref{fig:temporal_consistency}) against manually annotated 3D bounding boxes and semantic labels. We define two primary metrics:\begin{itemize}\item \textbf{Object Detection Recall:} Measured during the \textit{appearance interval} (when the object is present). A True Positive (TP) requires a 3D IoU $> 0.1$, centroid distance $< 0.3$m, and the correct semantic label.\item \textbf{Change Detection Recall:} Evaluates the system's ability to detect an object during its appearance and confirm its removal during the \textit{disappearance interval}. A TP requires detection in the former and zero false-positive detections in the latter.\end{itemize}As shown in Table~\ref{tab:object-change-detection}, DualMap \cite{jiang2025dualmap} achieves near-zero recall for object detection due to unstable 2D segmentation and inconsistent 3D box estimation, which causes most instances to be filtered out by its mapping module. While it maintains a nominal $\sim0.5$ recall for change detection, this is a mathematical artifact: the method ``succeeds" during the disappearance interval only because it failed to detect the object initially. Meanwhile, Khronos fails to produce consistent 3D estimation results (details shown in Supplementary) due to latent semantic inference bottlenecks that cause significant frame drops and degraded semantic mask quality. In contrast, SuperMap maintains high recall across both metrics, demonstrating robust instance detection.

\begin{table}[t]
\centering
\captionsetup{font=small}
\caption{\textbf{Spatial–Temporal Change Detection Performance.}}
\label{tab:object-change-detection}

\setlength{\tabcolsep}{3pt}
\footnotesize
\begin{tabular}{ll|ccc|ccc}
\toprule
\textbf{Method} & \textbf{Task}
& \multicolumn{3}{c|}{Appeared}
& \multicolumn{3}{c}{Disappeared} \\
\cmidrule(lr){3-5} \cmidrule(lr){6-8}
& & Buc. & Cart & Sign & Plant & Trash & Chair \\
\midrule
Khronos~\cite{schmid2024khronos}
& \multirow{3}{*}{Detect.}
& -- & -- & -- & -- & -- & -- \\
DualMap~\cite{jiang2025dualmap}
&
& 0.000 & 0.000 & 0.000 & 0.000 & 0.310 & 0.000 \\
\rowcolor{lightgray} \textbf{SuperMap (ours)}
&
& \textbf{1.000} & \textbf{0.262} & \textbf{0.583}
& \textbf{0.755} & \textbf{0.434} & \textbf{1.000} \\
\midrule
Khronos~\cite{schmid2024khronos}
& \multirow{3}{*}{Change}
& -- & -- & -- & -- & -- & -- \\
DualMap~\cite{jiang2025dualmap}
&
& 0.553 & 0.527 & 0.507
& 0.561 & 0.642 & 0.449 \\
\rowcolor{lightgray} \textbf{SuperMap (ours)}
&
& \textbf{1.000} & \textbf{0.622} & \textbf{0.790}
& \textbf{0.865} & \textbf{0.679} & \textbf{1.000} \\
\bottomrule
\label{tab:4}
\end{tabular}
\end{table}

\subsection{SuperMap System Ablation Study}

We evaluated the contribution of our individual sub-modules, specifically the 2D tracking, geometric consistency update, and Bayesian semantic fusion. Using a scene with 41 annotated objects, we quantified the performance of the spatio-temporal object map using precision, recall, and F1 scores.

\begin{table}[h]
    \centering
    \caption{\textbf{Performance Comparison of Ablation Study}}
    \label{tab:ablation}
    \begin{tabular}{lccc}
        \toprule
        \textbf{Configuration} & \textbf{Precision} & \textbf{Recall} & \textbf{F1} \\
        \midrule
        W/o 2D Tracker                   & 0.7787 & 0.4595 & 0.5780 \\
        W/o Semantic Fusion              & 0.7929 & 0.3870 & 0.5201 \\
        W/o Geometric Consistency Update & 0.8189 & 0.4448 & 0.5764 \\
        \rowcolor{lightgray} 
        \textbf{All (proposed)}             & \textbf{0.8677} & \textbf{0.4955} & \textbf{0.6308} \\
        \bottomrule
    \end{tabular}
\end{table}

As shown in Table \ref{tab:ablation}, the full system outperforms all baseline configurations across every metric, yielding the most accurate object map. This improvement stems from two primary factors. The 2D tracker maintains consistent identity association during significant viewpoint changes to prevent object fragmentation. Additionally, the integration of geometric and semantic fusion allows the system to remain resilient against momentary detection failures. By re-evaluating past inferences against new evidence over multiple frames, the map effectively self-corrects and filters out detection noise.


\begin{figure*}[ht]
    \centering
    \includegraphics[width=1\linewidth]{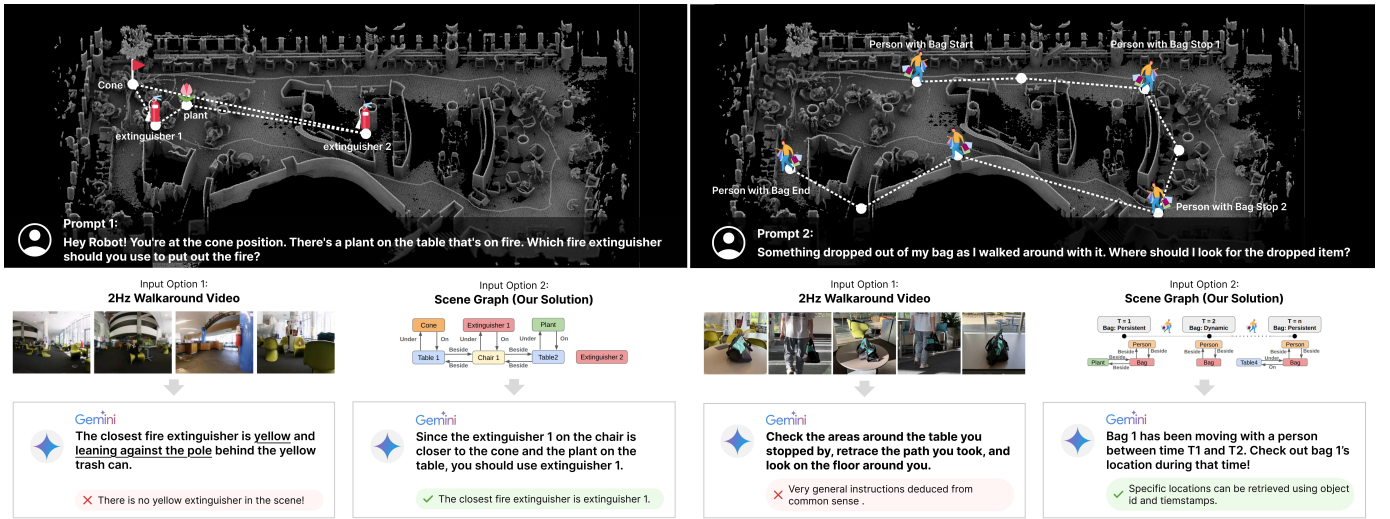}
    \captionsetup{font=small}
    \caption{\textbf{Comparison of Reasoning Results using Video Input and Scene-graph Input.} Compared to raw video, the scene-graph representation yields more reliable spatial reasoning, clearer temporal inference, and more accurate instance-level object retrieval.}
    \label{fig:spatio_temporal_reasoning_result}
    \vspace{-10.0pt}
\end{figure*}

\begin{figure*}[t]
    \centering
    \includegraphics[width=1.0\linewidth]{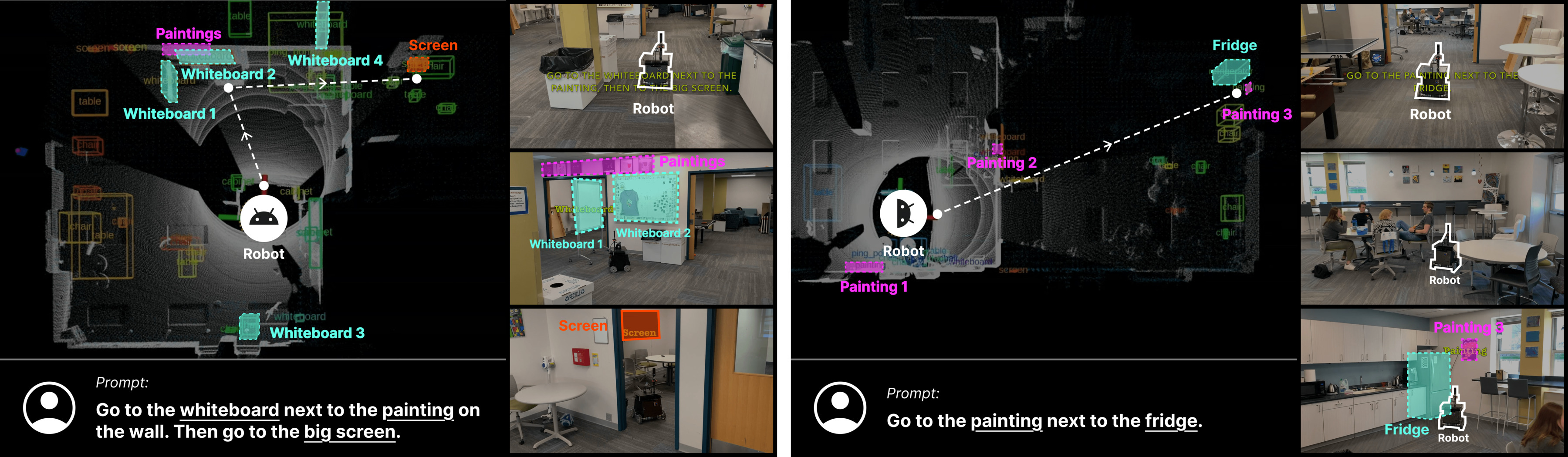}
    \captionsetup{font=small}
    \caption{\textbf{Online Visual–Language Navigation using Scene Graphs.} Our system enables robots to understand and reason about the spatial relationships in language instructions for VLN. See the supplementary video for more details. }
    \label{fig:VLN_navigation}
    \vspace{-5.0pt}
\end{figure*}




\subsection{Grounding Reasoning} \label{sec:reasoning}

We evaluate the 4D scene graph as a grounding engine for Large Vision-Language Models (VLMs) using Gemini 2.0 Flash~\cite{team2023gemini}. We compare reasoning performance between structured Scene Graph Input (serialized predicates) and raw Video Input (RGB sequences).
As shown in Fig.~\ref{fig:spatio_temporal_reasoning_result}, the 4D scene graph provides superior accuracy for complex queries:\begin{itemize}\item \textbf{Spatial Logic (Prompt 1):} 
Given a fire emergency scenario, the VLM utilizes the graph's metric edges to disambiguate a specific ``extinguisher" relative to a ``cone" and ``plant." While raw video suffers from perspective distortion and object confusion, the scene graph enables precise distance calculation and instance retrieval.\item \textbf{Temporal Logic (Prompt 2):} When tracing a ``bag's" past trajectory to find a dropped item, the VLM traverses the graph's temporal edges $E_T$. The structured history allows for filtered, timestamped retrieval of dynamic states. Conversely, video-based reasoning relies on unreliable ``common-sense" heuristics and suffers from hallucinations as temporal depth increases.\end{itemize}



\subsection{Visual Language Navigation}
\label{sec:vln}

We evaluate the downstream utility of SuperMap through zero-shot, language-guided navigation in previously unseen environments. Using the real-time 4D scene graph, the robot navigates reliably based solely on spatial relationships among 3D objects.

\paragraph{Onboard Grounding and Execution}
During autonomous exploration, the robot incrementally builds an instance-level semantic map. When given a natural language command (e.g., \textit{``Go to the whiteboard next to the painting, then go to the painting next to the fridge''}), the VLM interprets the spatial relations encoded in the 4D scene graph to resolve ambiguity and outputs precise 3D goal coordinates for the navigation stack.

\paragraph{Experimental Validation}
As shown in Fig.~\ref{fig:VLN_navigation}, Ours support navigation in visually ambiguous environments:
\begin{itemize}
    \item \textbf{Spatial ordering :} The robot selects the correct target among four visually identical whiteboards by reasoning over their relative spatial scene graph (left subfigure).
    \item \textbf{Relational retrieval:} The robot identifies the relationship between the fridge and the painting to reach the intended adjacent destination (right subfigure).
\end{itemize}

\subsection{Runtime and Memory} 
To assess the computational efficiency of our system, we measure the throughput of each module. The pose estimation module maintains a consistent 10 Hz output, while 2D instance segmentation operates at 1 Hz due to the heavy inference requirements of semantic segmentation. The remainder of the architecture performs 3D mapping and 4D scene graph updates at 3 Hz and 5 Hz, respectively.

\section{Conclusion} 
\label{sec:conclusion}
We present SuperMap, a real-time, open-vocabulary SLAM system that constructs and maintains an instance-level 4D semantic map. By integrating geometric constraints from SLAM with semantic features from 2D foundation models, SuperMap performs joint 2D-3D object tracking, instance validation, and change detection online. This unified pipeline ensures spatio-temporal consistency despite occlusions, partial observations, or dynamic environmental changes. The resulting 4D scene graph provides an explicit, continuously updated world model, empowering robots with long-term memory, robust spatio-temporal reasoning, and language-guided navigation.

\section{Limitation} 
\label{sec:limitation}
While SuperMap provides a robust framework for spatio-temporal mapping, its performance in tracking highly dynamic objects remains limited. Future iterations could address this by incorporating specialized tracking-by-detection modules or efficient segmentation-based tracking to improve temporal coherence. Furthermore, our current pipeline relies on a pre-defined list of object prompts for open-vocabulary 2D detection. Integrating an automated prompting mechanism based on open-world object discovery would significantly enhance the system's adaptability, allowing it to function in truly novel environments without prior semantic knowledge.

\bibliographystyle{plainnat}
\bibliography{references}
\clearpage

\setcounter{page}{1}
\setcounter{figure}{0}
\setcounter{table}{0}

\begingroup
\renewcommand{\thefootnote}{\fnsymbol{footnote}} 
\title{Supplementary: A Spatio-Temporal SLAM System for Visual-Language Navigation}
\author{
Shibo Zhao$^{\dagger}$,
Guofei Chen$^{\dagger}$,
Honghao Zhu,
Zhiheng Li,
Changwei Yao, \\
Nader Zantout,
Seungchan Kim,
Wenshan Wang,
Ji Zhang,
and Sebastian Scherer \\
The Robotics Institute, Carnegie Mellon University \\
{\tt\small \{guofeic, shiboz, basti\}@andrew.cmu.edu} \\
$^{\dagger}$Equal contribution
}
\date{} 
\maketitle
\endgroup

\section*{Introduction}

\begin{table*}[t]
\centering
\setlength{\tabcolsep}{4pt}
\caption{\textbf{3D instance segmentation on ScanNet with memory and runtime statistics.} TPF denotes mapping time per frame. Best results are highlighted as \colorbox{tabfirst}{\textbf{first}} and \colorbox{tabsecond}{second}.}
\resizebox{1.0\linewidth}{!}{%
\begin{tabular}{l|cc|cc|cc|cc|cc|c|c|c}
\toprule
\textbf{Method} & \multicolumn{2}{c|}{\textbf{Chair}} & \multicolumn{2}{c|}{\textbf{Window}} & \multicolumn{2}{c|}{\textbf{Refrigerator}} & \multicolumn{2}{c|}{\textbf{Sofa}} & \multicolumn{2}{c}{\textbf{Door}} & \textbf{Avg. Mem (MB)} & \textbf{Peak Mem (MB)} & \textbf{TPF (s)} \\
\cmidrule(lr){2-3} \cmidrule(lr){4-5} \cmidrule(lr){6-7} \cmidrule(lr){8-9} \cmidrule(lr){10-11}
& mAP$_{50}$ & mAP$_{25}$ & mAP$_{50}$ & mAP$_{25}$ & mAP$_{50}$ & mAP$_{25}$ & mAP$_{50}$ & mAP$_{25}$ & mAP$_{50}$ & mAP$_{25}$ \\
\midrule
HOV-SG \cite{werby2024hierarchical} & \cellcolor{tabsecond}4.58 & \cellcolor{tabsecond}4.73 & 0.00 & 0.00 & 0.00 & 0.00 & \cellcolor{tabsecond}30.00 & \cellcolor{tabsecond}31.25 & \cellcolor{tabsecond}9.70 & \cellcolor{tabsecond}10.40 & 8755.86 & 10226.66 & 8.623 \\
ConceptGraphs \cite{gu2024conceptgraphs} & 0.00 & 2.33 & 0.00 & 0.00 & 0.00 & 0.00 & 0.00 & 0.00 & 0.00 & 0.00 & \cellcolor{tabfirst}\textbf{859.86} & \cellcolor{tabfirst}\textbf{886.18} & \cellcolor{tabfirst}\textbf{0.092} \\
\textbf{SuperMap (Ours)} & \cellcolor{tabfirst}\textbf{63.76} & \cellcolor{tabfirst}\textbf{74.72} & \cellcolor{tabfirst}\textbf{42.20} & \cellcolor{tabfirst}\textbf{67.92} & \cellcolor{tabfirst}\textbf{62.50} & \cellcolor{tabfirst}\textbf{62.50} & \cellcolor{tabfirst}\textbf{33.35} & \cellcolor{tabfirst}\textbf{83.35} & \cellcolor{tabfirst}\textbf{10.00} & \cellcolor{tabfirst}\textbf{25.00} & \cellcolor{tabsecond}2030.09 & \cellcolor{tabsecond}2184.86 & \cellcolor{tabsecond}0.3604 \\
\bottomrule
\end{tabular}%
}
\label{tab:acc_mem_runtime}
\end{table*}

This supplementary material provides additional experimental results, qualitative analyses, implementation details, and clarifications that complement the main paper. Our goal is not to maximize isolated per-frame segmentation scores; instead, SuperMap is designed to improve robustness in dynamic environments and to enable long-horizon, object-centric scene understanding for downstream robotics tasks.

Most semantic SLAM systems~\cite{rosinol2020kimera,gu2024conceptgraphs} are evaluated using static perception metrics such as mIoU and mAP. While these metrics measure per-frame recognition quality, they do not capture a key requirement for real-world autonomy: \textit{long-horizon consistency} in dynamic scenes. In practice, robots should not only detect objects but also maintain stable instance identities as objects become occluded, move, leave the scene, or reappear. Achieving this requires reliable data association, temporal memory, and mechanisms to update the map as the environment evolves—capabilities that are largely unmeasured by standard segmentation benchmarks.

SuperMap addresses this limitation by maintaining object-level identities within a spatio-temporal map. As shown in the supplementary videos, the system can (i) remove objects that are no longer present, (ii) integrate newly observed objects, and (iii) re-associate previously seen objects under partial observations or moderate semantic drift. These properties are important for downstream tasks such as language-guided navigation, spatio-temporal reasoning, and long-term planning.

These observations motivate evaluation protocols that extend beyond static instance segmentation. To support this perspective, we provide additional qualitative analyses and real-robot deployment videos illustrating the system’s temporal behavior. As for the quantitative evaluation, it has been discussed in \tref{tab:4} in the main paper experiment section \ref{sec:experiment_settings} D.



\section*{Experiments}

\subsection{Qualitative Analysis of Object Tracking Ablations}
\label{sec:ablation_tracking}

Joint 2D--3D tracking and the consistency check are key components of SuperMap. We evaluate their contribution to (i) maintaining consistent instance identities over time, (ii) removing objects that disappear from the scene, and (iii) reducing projection-induced errors and other outliers.

\subsubsection{2D-Only Tracking vs.\ 2D--3D Tracking}
\begin{figure*}[t]
    \centering
    \includegraphics[width=\linewidth]{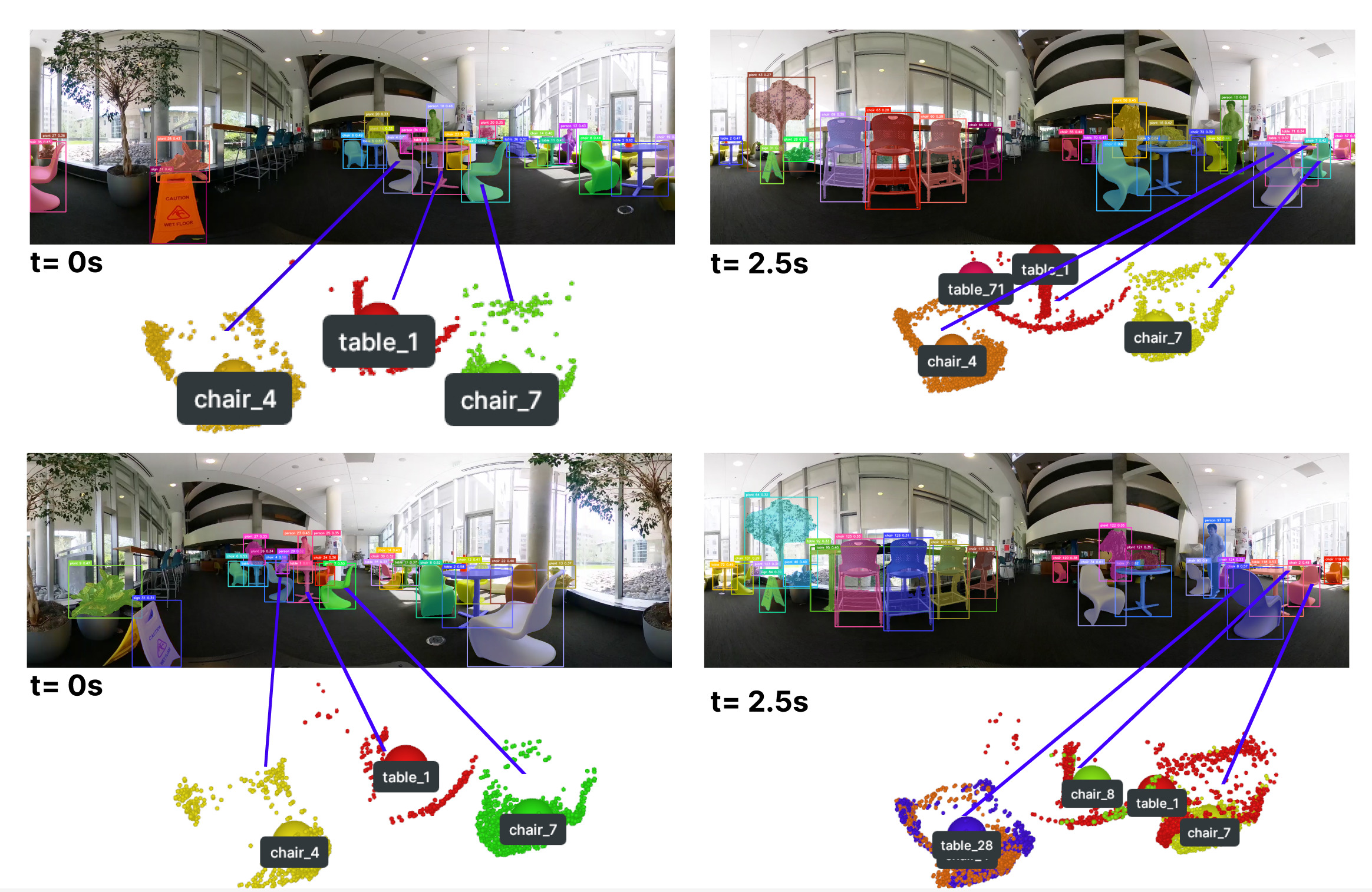}
    \caption{\textbf{Ablation on tracking under large viewpoint changes.} Our 2D–3D tracker (top) preserves consistent object tracks and yields a coherent map across significant viewpoint changes, whereas the 2D-only tracker (bottom) suffers from track fragmentation, resulting in incorrect object identities. Notably, the 2D-only tracker (bottom) exhibits clear ID switches between t = 0 s and t = 2.5 s.}
    \label{fig:abl_viewpoint_change}
\end{figure*}

\begin{figure*}
    \centering
    \includegraphics[width=\linewidth]{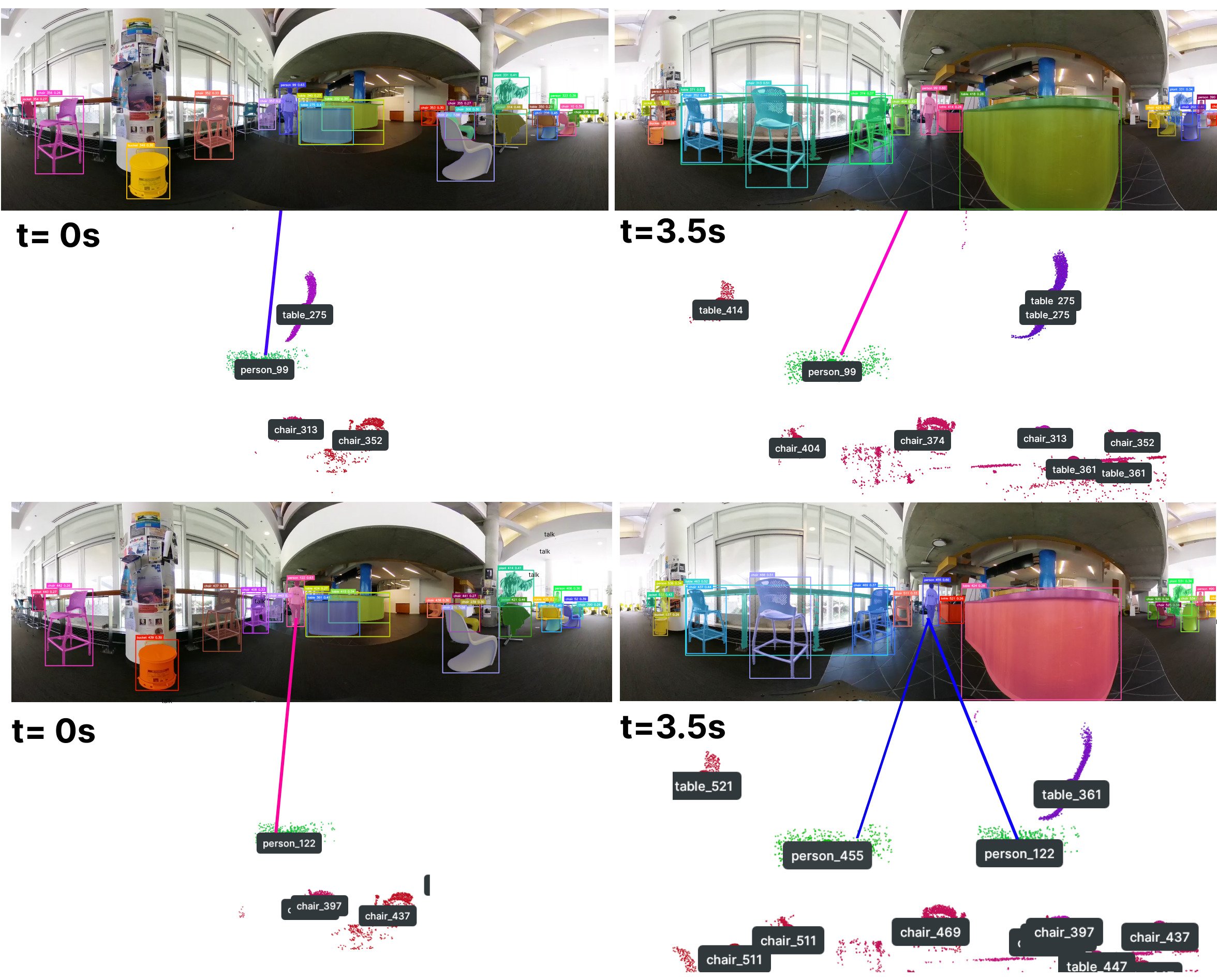}
  \caption{\textbf{Ablation on tracking dynamic objects.} The first row shows our 2D--3D tracking, while the second row shows 2D-only tracking. The 2D-only tracker frequently loses the target, resulting in the same person being represented as multiple objects in the map. In contrast, our 2D--3D tracker maintains a consistent identity over time.}
    \label{fig:abl_dyna_obj}
\end{figure*}

Figures~\ref{fig:abl_viewpoint_change} and~\ref{fig:abl_dyna_obj} compare a 2D-only ByteTrack\cite{zhang2022bytetrack} baseline with our 2D--3D variant, which refines 2D bounding-box states using estimated visibility and projections of 3D object centroids.

\begin{itemize}
    \item \textbf{Large viewpoint change.} In Fig.~\ref{fig:abl_viewpoint_change}, rapid camera motion causes the 2D-only tracker to lose tracks for nearby chairs and tables, leading to fragmented identities and degraded mapping. In contrast, the 2D--3D tracker preserves these identities by leveraging 3D cues, resulting in more temporally coherent instance IDs.
    \item \textbf{Dynamic objects.} In Fig.~\ref{fig:abl_dyna_obj}, a person moves across the robot's field of view. The 2D-only baseline repeatedly initializes new tracks for the same person, whereas our 2D--3D tracker maintains a single continuous ID, which is consistently reflected in the semantic map.
\end{itemize}

\subsubsection{Impact of the Consistency Check}
\label{sec:ablation_consistency}

\begin{figure*}
    \centering
    \includegraphics[width=0.6\linewidth]{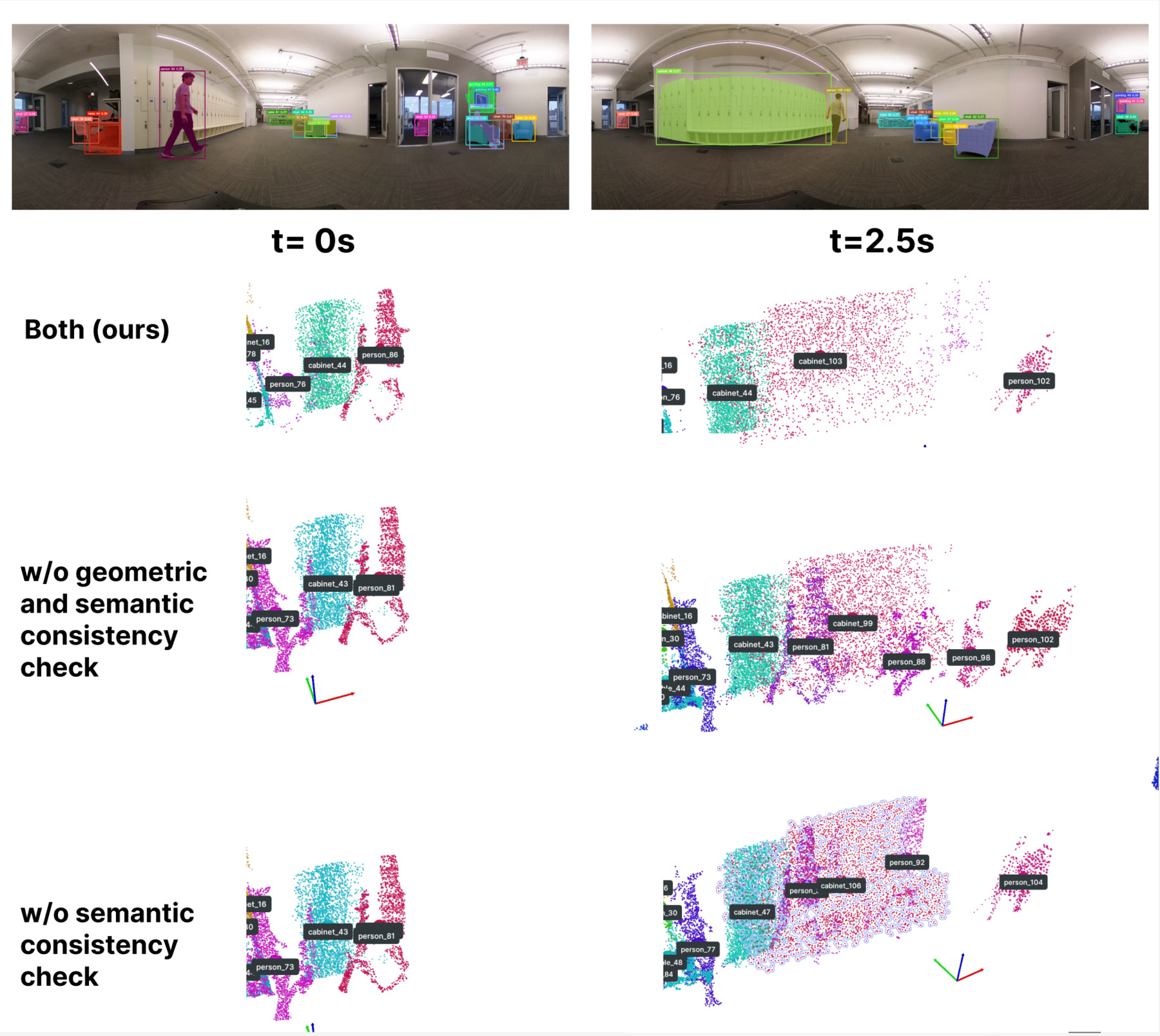}
   \caption{\textbf{Ablation of the consistency check.} Comparison of mapping results without consistency checks, with geometric-only checks, and with joint geometric+semantic checks (ours). Without consistency checks, stale points from dynamic objects persist in the map. Geometric checks remove some unsupported points but may leave projection artifacts. The joint geometric and semantic consistency check produces a cleaner and more up-to-date map.}

\label{fig:abl_consistency}
\end{figure*}

Figure~\ref{fig:abl_consistency} ablates the geometric and semantic consistency checks used to remove stale or spurious map points in a dynamic scene, where a person walks in front of a static cabinet:
\begin{enumerate}
    \item \textbf{No check.} Without consistency checking, points associated with the person persist after the person has moved away, leaving stale artifacts in the map.
    \item \textbf{Geometric only.} The geometric check removes unsupported points, but projection-induced outliers can still remain on the cabinet surface.
    \item \textbf{Geometric + semantic (ours).} Combining geometric and semantic checks removes both stale points and projection outliers, producing an up-to-date map.
\end{enumerate}
This ablation demonstrates that joint consistency checking is important for reliable mapping in dynamic environments.







\subsection{Qualitative Instance-Level Segmentation Results}
\label{sec:qual_instance_seg}
We present qualitative comparisons of instance-level segmentation results against recent state-of-the-art methods. Across representative sequences, SuperMap better preserves instance identities over time, whereas ConceptGraphs and HOV-SG more frequently fragment a single physical object into multiple instances. We also observe that the instance features used by ConceptGraphs and HOV-SG (based on CLIP embeddings) can become less discriminative over long sequences, which increases confusion between foreground objects and background regions.

\begin{figure*}[t]
    \centering
    \includegraphics[width=0.8\linewidth]{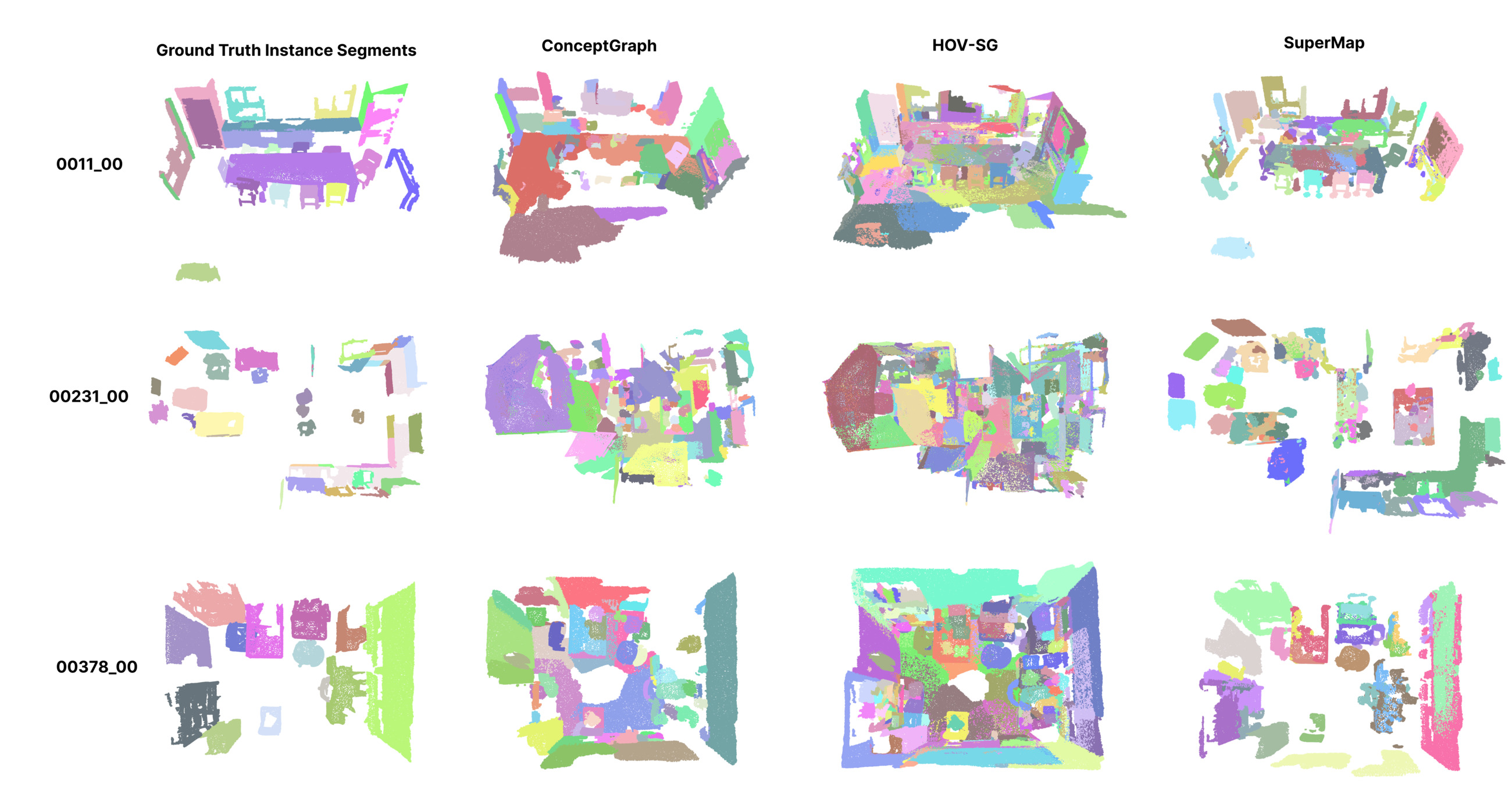}
    \caption{\textbf{Qualitative comparison of instance-level segmentation.} Background points are removed for clarity. SuperMap maintains more consistent instance identities and shows fewer confusions between objects and background compared to prior methods.}
    \label{fig:sota_comparison}
    \vspace{-5pt}
\end{figure*}

\subsection{Object-Level Mapping in Long-Term Dynamic Environments}
\label{sec:long_term_dynamic}

\begin{figure}[t]
    \centering
    \includegraphics[width=\linewidth]{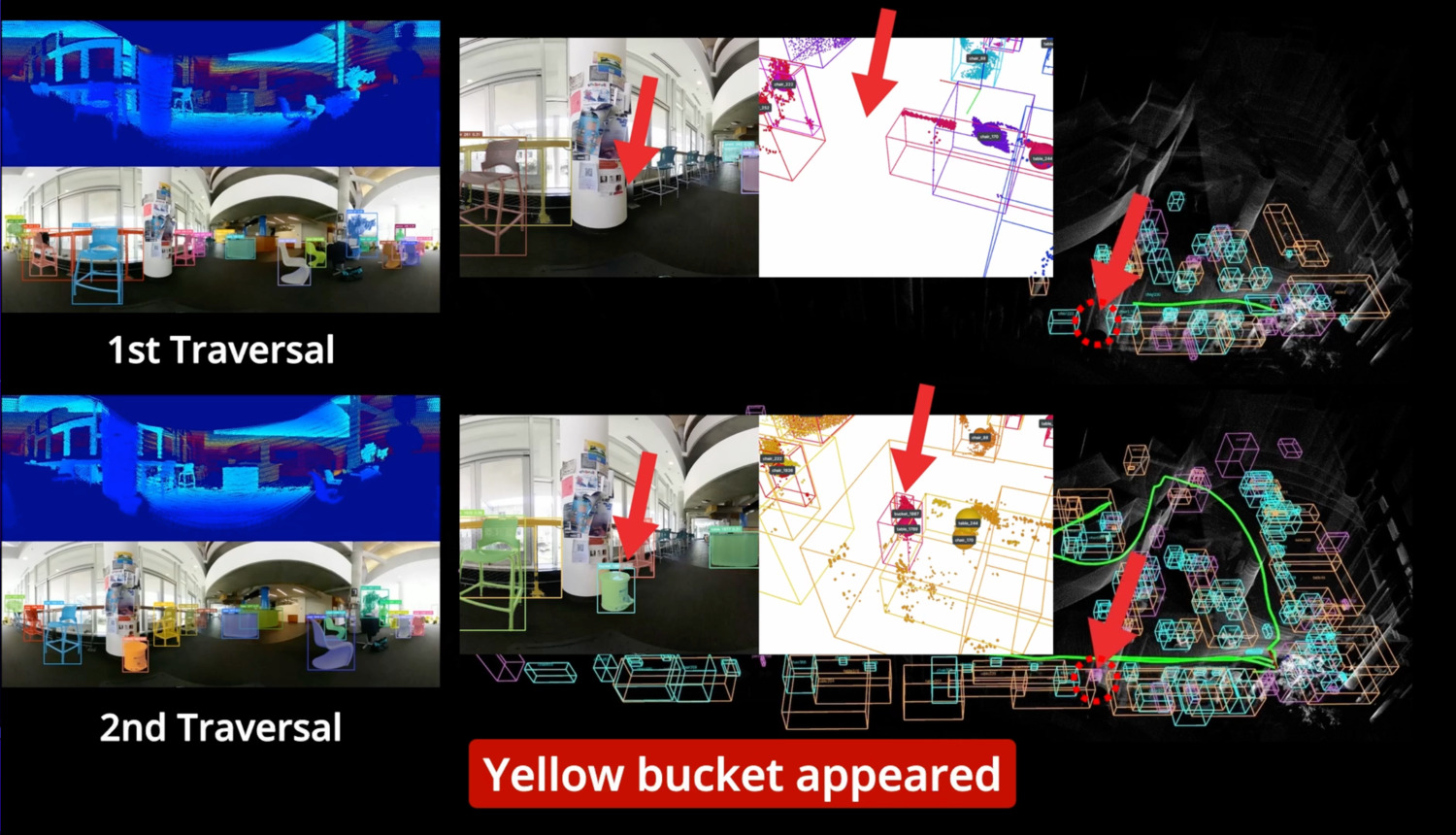}
    \caption{\textbf{Newly appeared object (yellow bucket).} When revisiting the environment, SuperMap detects the newly introduced bucket and incorporates it into the object-level map.}
    \label{fig:bucket_appear}
    \vspace{-4pt}
\end{figure}

\begin{figure}[t]
    \centering
    \includegraphics[width=\linewidth]{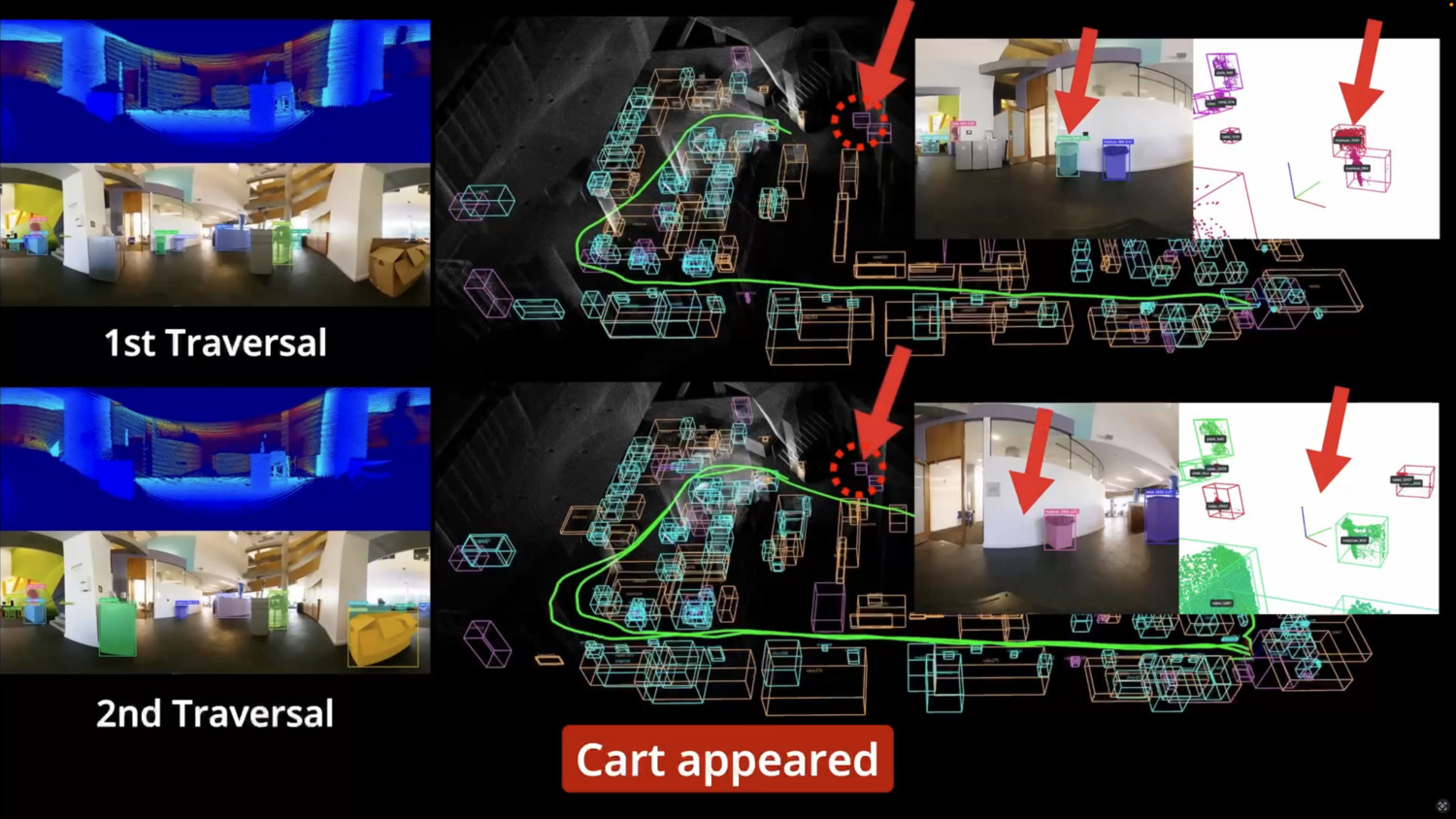}
    \caption{\textbf{Newly appeared object (cart).} The system recognizes the cart as a new instance and updates the map accordingly.}
    \label{fig:cart_appear}
    \vspace{-4pt}
\end{figure}

\begin{figure}[t]
    \centering
    \includegraphics[width=\linewidth]{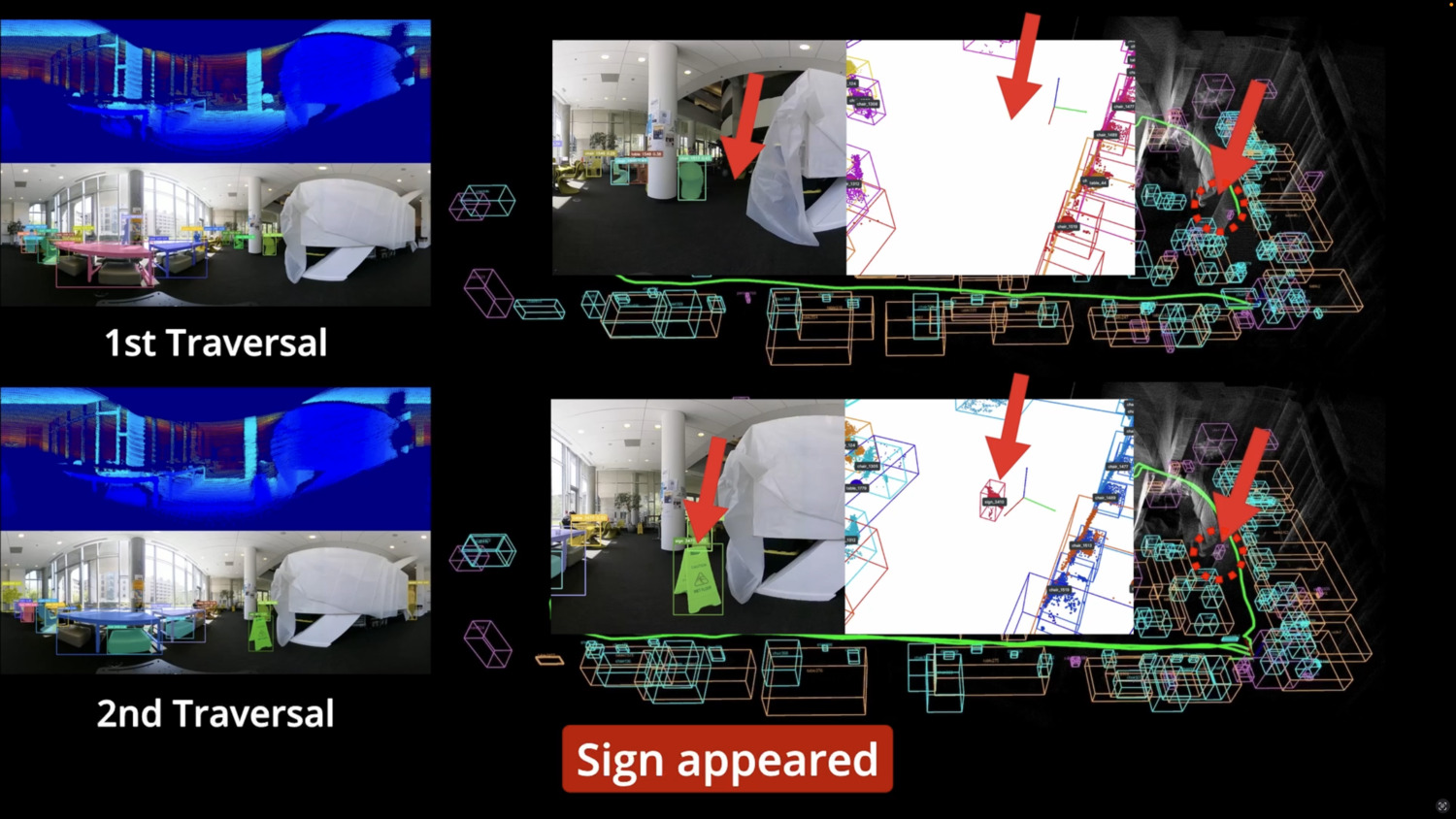}
    \caption{\textbf{Newly appeared object (safety sign).} SuperMap adds the newly observed sign to the spatio-temporal map.}
    \label{fig:sign_appear}
    \vspace{-4pt}
\end{figure}

\begin{figure}[t]
    \centering
    \includegraphics[width=\linewidth]{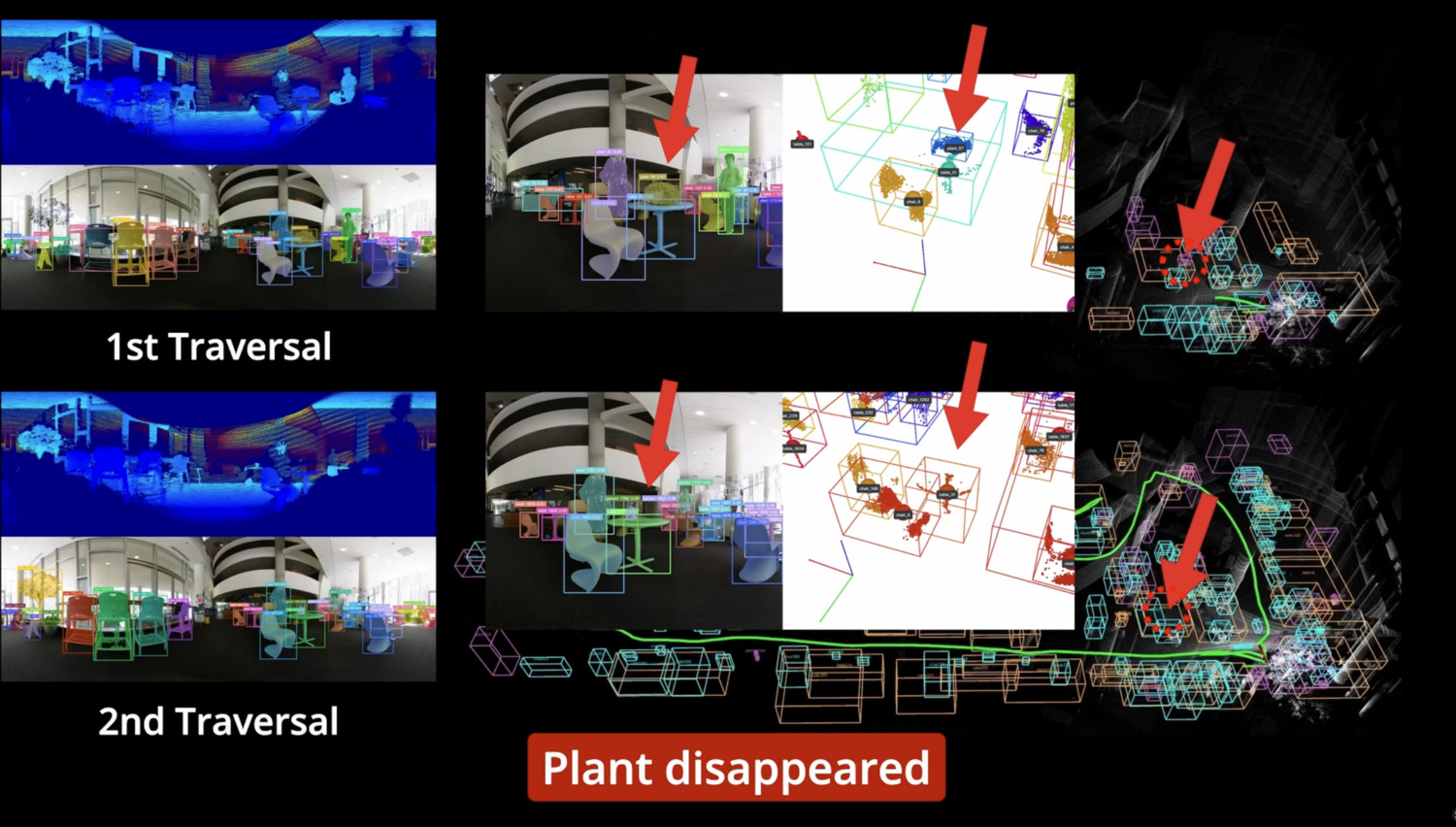}
    \caption{\textbf{Disappeared object (plant on table).} The system removes the plant from the map after it is no longer observed.}
    \label{fig:plant_disappear}
    \vspace{-4pt}
\end{figure}

\begin{figure}[t]
    \centering
    \includegraphics[width=\linewidth]{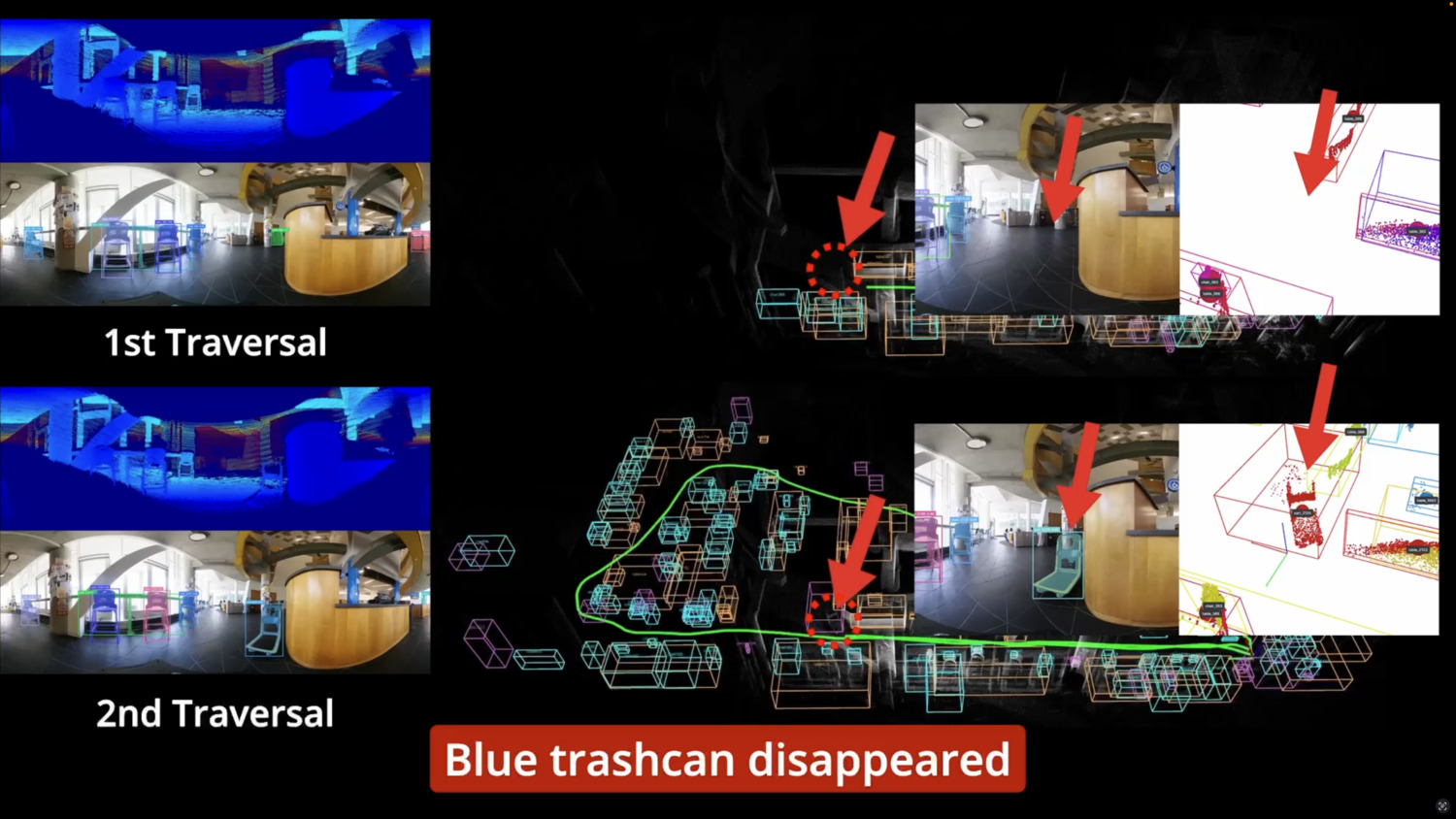}
    \caption{\textbf{Disappeared object (blue trash can).} SuperMap updates the map to reflect the removal of the trash can.}
    \label{fig:trashcan_disappear}
    \vspace{-4pt}
\end{figure}

\begin{figure}[t]
    \centering
    \includegraphics[width=\linewidth]{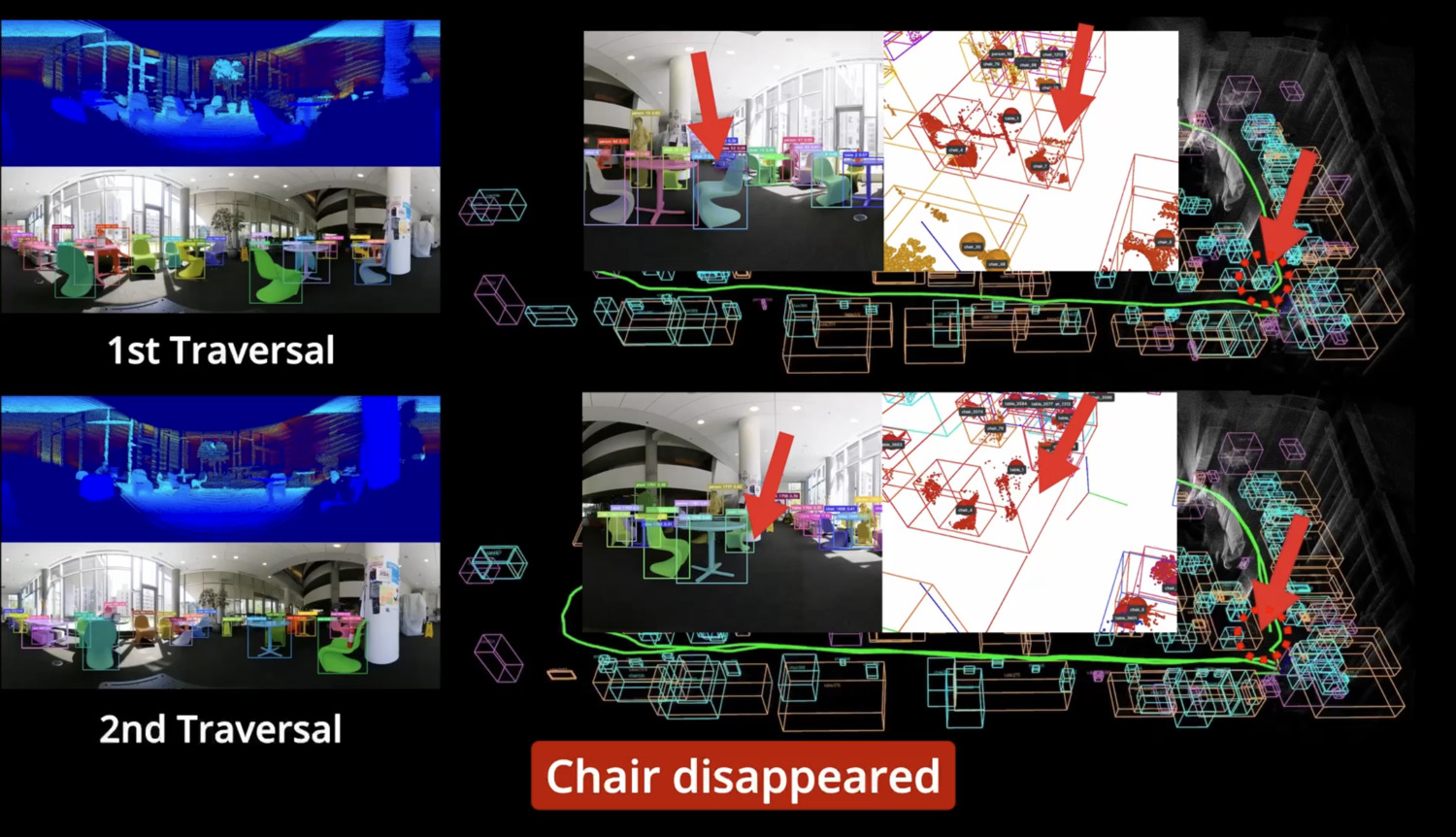}
    \caption{\textbf{Disappeared object (chair).} The system correctly removes the chair from the object-level map once it is absent.}
    \label{fig:chair_disappear}
    \vspace{-4pt}
\end{figure}

Figures~\ref{fig:bucket_appear}--\ref{fig:chair_disappear} illustrate long-term scene changes across two traversals of the same environment. We highlight three newly appeared objects (a yellow bucket, a cart, and a safety sign) and three disappeared objects (a plant on a table, a blue trash can, and a chair). These examples demonstrate that SuperMap can both identify object-level changes and update the spatio-temporal map accordingly. Additional qualitative results are provided in the supplementary video.


\subsection{Runtime Details of Each Component}
\label{sec:runtime_details}

For reproducibility, we report the update rates of each module in our system:
\begin{itemize}
    \item Pose estimation: 10~Hz
    \item Dense 3D mapping and 3D bounding box update: 3~Hz
    \item 2D instance segmentation: 1~Hz
    \item 4D scene graph update: 5~Hz
\end{itemize}

Memory usage and mapping-time statistics are summarized in \tref{tab:acc_mem_runtime}.
As shown in Table~\ref{tab:acc_mem_runtime}, SuperMap achieves substantially stronger instance-level performance across categories while maintaining moderate runtime and memory usage. Although our method does not achieve the lowest memory footprint, this is expected because SuperMap maintains persistent object-level representations, temporal associations, and a 4D scene graph to support long-horizon consistency. These design choices intentionally trade minimal memory usage for temporally consistent mapping and richer scene understanding. In practice, the memory consumption remains within the capacity of standard robotic computing platforms and enables capabilities that lightweight per-frame methods do not provide.

ConceptGraph incorporates CLIP features to support open-vocabulary object querying. However, the queried semantic features are typically sparse in the reconstructed map. As a result, while target objects can be retrieved, the resulting spatial localization remains coarse, and the precision falls significantly below standard detection benchmarks such as mAP@50 and mAP@25.

\section{Method}

We present the spatio-temporal object update procedure used in our semantic SLAM system. The algorithm processes a sequence of observations comprising detected boxes $B_{\text{det}}$, an RGB image $I$, a depth map $D_I$, and pose information $\text{Odom}$ (odometry or estimated camera pose). We use a multi-object tracker $\mathcal{T}$ (Kalman-filter based, following ByteTrack~\cite{zhang2022bytetrack}) to maintain temporal identity consistency across frames.

\begin{algorithm}[t]
\caption{Instance Layer: Spatio-Temporal Instance Association}
\label{alg:val_track}
\begin{algorithmic}[1]
\Require Observation stream $S = \{(B_{\text{det}}, I, D_I, \text{Odom})\}$; tracker $\mathcal{T}$
\Ensure Updated map $M$

\State $M \gets \texttt{InitMap()}$; \quad $\mathcal{T}\!\gets\!\texttt{InitTracker()}$

\ForAll{$(B_{\text{det}}, I, D_I, \text{Odom}) \in S$}

    \Statex \textbf{(1) Gather local map context}
    \State $O_{\text{nbr}} \gets \texttt{GetNeighborObjects}(M, \text{Odom})$

    \Statex \textbf{(2) Geometric consistency check (Eq.~\ref{eq:consistency})}
    \State $(O_{\text{front}}, O_{\text{on}}, O_{\text{behind}}) \gets \texttt{CompareDepth}(D_I, O_{\text{nbr}}, \text{Odom})$
    \State $O_{\text{neg}} \gets O_{\text{front}}$ \Comment{inconsistent depth}

    \Statex \textbf{(3) Track update and detection association}
    \State $T_{\text{alive}} \gets \mathcal{T}.\texttt{GetAliveTracks()}$
    \ForAll{$O \in O_{\text{on}}$}
        \If{$O.\texttt{id} \in T_{\text{alive}}$}
            \State $\mathcal{T}.\texttt{UpdateFromMap}(O)$ \Comment{compensate track state using 3D cues}
        \EndIf
    \EndFor
    \State $T_{\text{trk}} \gets \mathcal{T}.\texttt{Associate}(T_{\text{alive}}, B_{\text{det}})$ \Comment{ByteTrack~\cite{zhang2022bytetrack}}

    \Statex \textbf{(4) Instance mask and 3D reconstruction}
    \State $M_{\text{mask}} \gets \texttt{GenerateMasks}(I, T_{\text{trk}})$ \Comment{SAM2~\cite{ravi2024sam}}
    \State $O_{\text{add}} \gets \texttt{Unproject}(D_I, M_{\text{mask}}, \text{Odom})$

    \Statex \textbf{(5) Semantic consistency check and map update}
    \ForAll{$O \in O_{\text{on}}$}
        \If{$O.\texttt{id} \notin T_{\text{trk}}$}
            \State $O_{\text{neg}} \gets O_{\text{neg}} \cup \{O\}$ \Comment{semantic inconsistency}
        \EndIf
    \EndFor

    \State $O_{\text{nbr}} \gets \texttt{ObjNegate}(O_{\text{nbr}}, O_{\text{neg}})$
    \State $O_{\text{nbr}} \gets \texttt{ObjMerge}(O_{\text{nbr}}, O_{\text{add}})$
    \State $M \gets \texttt{MapMerge}(M, O_{\text{nbr}})$
\EndFor

\end{algorithmic}
\end{algorithm}

\subsection{2D-3D Consistency Check}

To update objects in the map, we perform two key checks:

\begin{itemize}
    \item Whether the object's voxels remain occupied.
    \item Whether the occupied and visible voxels are supported by the latest semantic inference from the image.
\end{itemize}

The geometric consistency check is implemented in lines 2–5, while the semantic consistency is in lines 6–19.

\subsection{3D-aware Instance Tracking}

To enhance the matching between new bounding boxes and existing map objects, we introduce a 3D-aware instance tracking mechanism.

Our tracking algorithm builds on Algorithm 1 from ByteTrack \cite{zhang2022bytetrack}, preserving the original 2D state representation and motion model for tracklets. In addition, we incorporate geometric cues to update the internal states of the tracklets.

\end{document}